\documentclass[USenglish,oneside,twocolumn]{article}

\usepackage[utf8]{inputenc}%(only for the pdftex engine)
\usepackage[big]{dgruyter_NEW}
 
\DOI{foobar}

\cclogo{\includegraphics{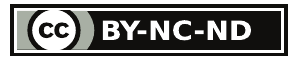}}

\usepackage{amsmath}
\usepackage{amssymb}
\usepackage{mathtools}
\usepackage{resizegather}
\usepackage{thm-restate}
\usepackage{xspace}
\usepackage{mathrsfs}
\usepackage{ bbold }

\usepackage[utf8]{inputenc} % allow utf-8 input
\usepackage[T1]{fontenc}    % use 8-bit T1 fonts
\usepackage{hyperref}       % hyperlinks
\usepackage{url}            % simple URL typesetting
\usepackage{booktabs}       % professional-quality tables
\usepackage{amsfonts}       % blackboard math symbols
\usepackage{nicefrac}       % compact symbols for 1/2, etc.
\usepackage{microtype}      % microtypography
\usepackage{float}          % floating algorithms

\usepackage{lipsum}                     % Dummytext
\usepackage{xargs}                      % Use more than one optional parameter in a new commands
\usepackage[pdftex,dvipsnames,table]{xcolor}  % Coloured text etc.
\usepackage[colorinlistoftodos,prependcaption,textsize=small]{todonotes}
\newcommandx{\unsure}[2][1=]{\todo[linecolor=red,backgroundcolor=red!25,bordercolor=red,#1]{#2}}
\newcommandx{\change}[2][1=]{\todo[linecolor=blue,backgroundcolor=blue!25,bordercolor=blue,#1]{#2}}
\newcommandx{\info}[2][1=]{\todo[linecolor=OliveGreen,backgroundcolor=OliveGreen!25,bordercolor=OliveGreen,#1]{#2}}
\newcommandx{\improvement}[2][1=]{\todo[linecolor=Plum,backgroundcolor=Plum!25,bordercolor=Plum,#1]{#2}}

\usepackage{caption}
\usepackage[noend,linesnumbered,lined,boxed]{algorithm2e}
\usepackage{enumitem}
\usepackage{wrapfig}
\usepackage{varwidth} % for the varwidth minipage environment
\newcolumntype{M}{>{\begin{varwidth}{1cm}}l<{\end{varwidth}}}
\usepackage{wrapfig}
\usepackage{listings}

\newcommand{\Z}{\mathbb{Z}}

\newcommand{\xor}{\oplus}

\newcommand{\kb}{\mathsf{k_{j}}}
\newcommand{\key}{\mathsf{k}}
\newcommand{\sep}{\;||\;}
\newcommand{\bool}{j}

\newcommand{\CWi}{CW^{(i)}}

\newcommand{\zerone}{\{0, 1\}}

\newcommand{\sampled}{\overset{{}_\$}{\leftarrow}}
\newcommand{\shared}[1]{[\![ #1 ]\!]}
\SetAlCapSkip{.5em}
\SetKwInput{KwInitialisation}{Initialisation}%
\SetKwComment{nol}{}{}

\usepackage{amsthm}

\theoremstyle{definition}
\newtheorem{definition}{Definition}[section]
\newtheorem{claim}{Claim}[section]

\begin{document}
 
  %\author*[1]{Anonymous authors}

  \author*[1]{Théo Ryffel}

  \author[2]{Pierre Tholoniat}

  \author[3]{David Pointcheval}

  \author[4]{Francis Bach}
  
  %\affil[1]{E-mail: anonymous@...}
  \affil[1]{INRIA, Département d’informatique de l’ENS, ENS, CNRS, PSL University, Arkhn, Paris, France, E-mail: theo.ryffel@ens.fr}

  \affil[2]{Columbia University, New York, USA}

  \affil[3]{Département d’informatique de l’ENS, ENS, CNRS, PSL University, INRIA, Paris, France}

  \affil[4]{INRIA, Département d’informatique de l’ENS, ENS, CNRS, PSL University, Paris, France}

  \title{\huge AriaNN: Low-Interaction Privacy-Preserving Deep Learning via Function Secret Sharing}

  \runningtitle{AriaNN: Low-Interaction Privacy-Preserving Deep Learning via Function Secret Sharing}

  %\subtitle{...}

  \begin{abstract}{
We propose \textsc{\large A}\textsc{riaNN}, a low-interaction privacy-preserving framework for private neural network training and inference on sensitive data. \\
\indent Our semi-honest 2-party computation protocol (with a trusted dealer) leverages function secret sharing, a recent lightweight cryptographic protocol that allows us to achieve an efficient online phase. We design optimized primitives for the building blocks of neural networks such as ReLU, MaxPool and BatchNorm. For instance, we perform private comparison for ReLU operations with a single message of the size of the input during the online phase, and with preprocessing keys close to $4\times$ smaller than previous work. Last, we propose an extension to support $n$-party private federated learning.\\
\indent We implement our framework as an extensible system on top of PyTorch that leverages CPU and GPU hardware acceleration for cryptographic and machine learning operations. 
We evaluate our end-to-end system for private inference between distant servers on standard neural networks such as AlexNet, VGG16 or ResNet18, and for private training on smaller networks like LeNet.
% We evaluate our end-to-end system for private inference and training on standard neural networks such as AlexNet, VGG16 or ResNet18 between distant servers. 
We show that computation rather than communication is the main bottleneck and that using GPUs together with reduced key size is a promising solution to overcome this barrier.
}
\end{abstract}
%  \classification[PACS]{}
 % \communicated{...}
 % \dedication{...}

\journalname{Proceedings on Privacy Enhancing Technologies}
\DOI{Editor to enter DOI}
\startpage{1}
%\received{..}
%\revised{..}
%\accepted{..}

\journalyear{2022}
%\journalvolume{..}
%\journalissue{..}

\maketitle

\section{Introduction}
%\subsection{Context}

The massive improvements of cryptography techniques for secure computation over sensitive data \cite{damgaard2012multiparty, chillotti2016faster, keller2018overdrive} have spurred the development of the field of privacy-preserving machine learning \cite{shokri2015privacy, al2019privacy}. Privacy-preserving techniques have become practical for concrete use cases, thus encouraging public authorities to use them to protect citizens' data especially in healthcare applications \cite{kaur2018efficient, dugan2016survey, reichert2020privacy}.

However, tools are lacking to provide end-to-end solutions for institutions that have little expertise in cryptography while facing critical data privacy challenges. A striking example is hospitals, which handle large amounts of data while having relatively constrained technical teams. Secure multi-party computation (SMPC) is a promising technique that can be efficiently integrated into machine learning workflows to ensure data and model privacy, while allowing multiple parties or institutions to participate in a joint project. In particular, SMPC provides intrinsic shared governance: because data is shared, none of the parties can decide alone to reconstruct it. 
%This is particularly suited for collaborations between institutions willing to share ownership on a trained model. 

\vspace*{.111cm}

\noindent \textbf{Use case.}
The main use case driving our work is the collaboration between a healthcare institution and an AI company. The healthcare institution, an hospital for example, acts as the data owner and the AI company as the model owner. The collaboration consists of either training the model with labelled data or using a pre-trained model to analyze unlabelled data. Training can possibly involve several data owners, as detailed in Section \ref{section:n_party}. Since the model can be a sensitive asset (in terms of intellectual property, strategic asset or regulatory and privacy issues), it cannot be trained directly on the data owner(s) machines using techniques like federated learning \cite{konevcny2016federated, bonawitz2019towards}: it could be stolen or reverse-engineered \cite{hitaj2017deep, fredrikson2015model}.

We will assume that the parties involved in the computation are located in different regions, and that they can communicate large amounts of information over the network with a reasonable latency (70ms for example). This corresponds to the \emph{Wide Area Network (WAN)} setting, as opposed to the \emph{Local Area Network (LAN)} setting where parties are typically located in the same data center and communicate with low latency (typically <$1$ms). Second, parties are \emph{honest-but-curious}, \cite[Chapter~7.2.2]{goldreich2009foundations} and care about their reputation. Hence, they have little incentive to deviate from the original protocol, but they will use any information available in their own interest.  

\vspace*{.111cm}

\noindent \textbf{Contributions.}
By leveraging function secret sharing (FSS) \cite{boyle2015function, boyle2016function}, we propose a low-interaction framework for private deep learning which drastically reduces communication to a single round for basic machine learning operations, and achieves the first private evaluation benchmark on ResNet18 using GPUs.
%for basic machine learning operations such as matrix multiplication or comparison
%By drastically reducing the number of interactions compared to existing solutions, this framework makes training of a private model across $n$ distant parties practical.

\begin{itemize}[itemsep=2mm,nolistsep,leftmargin=2em]

  \item[\textbullet] We improve upon existing work of \cite{boyle2016function} on function secret sharing to design compact and ready-to-implement algorithms for tensor private comparison, which is a building block for neural networks and can be run with a single round of communication. In particular, we significantly reduce the key size from roughly $n(4 \lambda + n)$ to $n(\lambda + 2n)$, which is a crucial parameter as the computation time is linear in the key size.

  \item[\textbullet] We show how function secret sharing can be used in machine learning and provide privacy-preserving implementations of classical layers, including ReLU, MaxPool and BatchNorm, to allow secure evaluation and training of arbitrary models on private data.

  % \item We show how this framework adapts to the case where $n$ parties contribute their data.

  \item[\textbullet] Last, we provide a GPU implementation and a hardware-accelerated CPU implementation of our private comparison protocol\footnote{The code is available at \href{https://github.com/LaRiffle/AriaNN}{github.com/LaRiffle/AriaNN}.}. As AriaNN is built over PyTorch for other tensor operations, it can run either completely on the GPU or on the CPU. We show its practicality both in LAN and WAN settings by running private inference on CIFAR-10 and Tiny Imagenet with models such as AlexNet \cite{krizhevsky2012alexnet}, VGG16 \cite{simonyan2014vgg} and ResNet18 \cite{he2016resnet}, and private training on MNIST using models like LeNet.
\end{itemize}

%All code and implementations will be open-sourced on GitHub will be integrated in an open-source library and will be made public on GitHub. %to the PySyft library and can be found online at \url{github.com/OpenMined/PySyft}.

\vspace*{.211cm}

\noindent \textbf{Related work.}
Related work in privacy-preserving machine learning encompasses SMPC and fully homomorphic encryption (FHE) techniques.%, which have been used for secure evaluation or training of models.

FHE only needs a single round of interaction but does not support efficient non-linearities. For example, nGraph-HE  \cite{boemer2019ngraph} and its extensions \cite{boemer2019ngraph2} build on the SEAL library \cite{sealcrypto} and provide a framework for secure evaluation that greatly improves on the CryptoNet seminal work~\cite{gilad2016cryptonets}, but it resorts to polynomials (like the square) for activation functions.
%does not remove the constraint from the original paper that activation functions should be polynomials (like the square) and not widely-used functions like ReLU.

SMPC frameworks usually provide faster implementations using lightweight cryptography. MiniONN \cite{liu2017oblivious}, DeepSecure \cite{rouhani2018deepsecure} and XONN \cite{riazi2019xonn} use optimized garbled circuits \cite{yao1986generate} that allow very few communication rounds, but they do not support training and alter the neural network structure to speed up execution. Other frameworks such as ShareMind \cite{bogdanov2008sharemind}, SecureML \cite{mohassel2017secureml}, SecureNN \cite{wagh2018securenn}, QUOTIENT \cite{agrawal2019quotient} or more recently FALCON \cite{wagh2020falcon} rely on additive secret sharing and allow secure model evaluation and training. They use simpler and more efficient primitives, but require a large number of rounds of communication, such as 11 in  \cite{wagh2018securenn} or $5 + \log_2(n)$ in \cite{wagh2020falcon} (typically 10 with $n = 32$) for ReLU. ABY \cite{demmler2015aby}, Chameleon \cite{riazi2018chameleon} and more recently $\textrm{ABY}^3$ \cite{mohassel2018aby3}, CrypTFlow \cite{kumar2020cryptflow} and \cite{fantasticfour} mix garbled circuits, additive or binary secret sharing based on what is most efficient for the operations considered. However, conversion between those can be expensive and they do not support training except $\textrm{ABY}^3$. There is a current line of work including BLAZE \cite{patra2020blaze}, Trident \cite{chaudhari2020trident} and FLASH \cite{byali2020flash} which improves over $\textrm{ABY}^3$ to reduce communication overheads: BLAZE and Trident achieve for example 4 rounds of communication for ReLU.

%\textit{Last, except \cite{wagh2018securenn}, \cite{wagh2020falcon} and \cite{mohassel2018aby3}, they are only considered secure in the honest-but-curious model.}\todo{TR: rm this if we don't do better} 

Last, works like Gazelle \cite{juvekar2018gazelle} combine FHE and SMPC to make the most of both, but conversion can also be costly.%\change{FB: say why we don't compare explicitly in simulations to any of these. TR: actually we can compare favourably to Gazelle}

Works on trusted execution environments are left out of the scope of this article as they require access to dedicated and expensive hardware \cite{hunt2018chiron}. 
% Data owners which cannot afford these secure enclaves might be reluctant to use a cloud service and to send their data. 
% Differential privacy is also out of our scope, even it would be useful if the model needs to be disclosed publicly and has been trained on sensitive data. Integrating it into our framework would be a natural continuation of this work. 

A concurrent work from Boyle et al. \cite{boylefunction2020} was made public shortly after ours. Their approach also provides improvement over previous algorithms for private comparison using function secret sharing, and their implementation results in the same number of rounds than ours and similar key size (approximately $n(\lambda + n)$, where $n$ is the number of bits to encode the value, it accounts for correctness and is typically set to 32, and $\lambda$ is the security parameter and usually equals 128). However, \cite{boylefunction2020} is not intended for machine learning: they only provide an implementation of ReLU, but not of MaxPool, BatchNorm, Argmax or other classic machine learning components. In addition, as they do not provide experimental benchmarks or an implementation of their private comparison, we are not able to compare it to ours in our private ML framework. They avoid the negligible error rate that we study in Section \ref{sec:failure_rate}, which has no impact in the context of machine learning as we show.

% Interestingly, \cite{boylefunction2020} proposes a way to combine the 2 invocations of \cite{boyle2019secure} in a single evaluation while keeping the key size roughly the same. This idea could be useful in scenarios where the error rate wouldn't be acceptable, like when comparing huge numbers.

%$n(\lambda + 2n + 4) + \lambda + 2n$ bits. This allows for faster transmission of keys over the network to the parties doing the evaluation. 
% We note that \cite{boylefunction2020} achieves similar key size for comparison, with $n(\lambda + n + 2) + \lambda + 2n$ bits.

\section{Background}\label{section:background}
\textbf{Notations.}
All values are encoded on $n$ bits and live in $\Z_{2^n}$. The bit decomposition of any element $x$ of $\Z_{2^n}$ into a bit string of $\zerone^n$ is a bijection between $\Z_{2^n}$ and $\zerone^n$. Therefore, bit strings generated by a pseudo random generator $G$ are implicitly mapped to $\Z_{2^n}$. In addition, we interpret the most significant bit as a sign bit to map them in $[-2^{n-1},2^{n-1}-1]$, notably in Algorithms \ref{algo:key_gen_eq}, \ref{algo:eval_eq}, \ref{algo:key_gen_comp}, \ref{algo:eval_comp}, \ref{algo:protocol_comp}, where the modulo operation makes the conversion between n bit strings and signed integers explicit.

The notation $\shared{x}$ denotes 2-party additive secret sharing of $x$, i.e., $\shared{x} = (\shared{x}_0, \shared{x}_1)$ where the shares $\shared{x}_j$ are random in $\Z_{2^n}$, are held by distinct parties and verify $x = \shared{x}_0 + \shared{x}_1 \bmod 2^n$. In return, $x[i]$ refers to the $i$-th bit of $x$. 
The comparison operator $\le$ is taken over the natural embedding of $\Z_{2^n}$ into $\Z$.
%inally, $n$-bit signed integers in $[-2^{n-1},2^{n-1}-1]$ can be mapped to $\Z_{2^n}$ with the $\mod 2^n$ operator, and we can interpret the most significant bit (MSB) as a sign bit.
% \todo[inline]{PT: add more details about the mapping here (instead of inside the example that follows)? Explain what ``wrapping around'' means and how we define $\le$. TR: Yes that would be better indeed!}

%\info{FB: not clear what $\shared{x}$ formally is. It belongs to which space?} %\info{FB: relate $\Z_{2^n}$ to number of bits}

%We use the notation $\shared{x}$ \todo{FB: not clear what $\shared{x}$ formally is. It belongs to which space?} to denote additive secret sharing of an element $x$ in $\Z_{2^n}$. \todo{FB: relate $\Z_{2^n}$ to number of bits} The shares $\shared{x}_i$ of $\shared{x} = (\shared{x}_0, \shared{x}_1)$ are random, each is held by a different party and they verify $x = \shared{x}_0 + \shared{x}_1 \bmod 2^n$. In return, $x[i]$ refers to the $i$-th bit of $x$.

\subsection{Function Secret Sharing} \label{sec:background:fss}
Unlike classical data secret sharing, where a shared input $\shared{x}$ is applied on a public $f$, function secret sharing applies a public input $x$ on a private shared function $\shared{f}$. Shares or \emph{keys} $(\shared{f}_0, \shared{f}_1)$ of a function $f$ satisfy $f(x) = \shared{f}_0(x) + \shared{f}_1(x) \bmod 2^n$ and they can be provided by a semi-trusted dealer. Both approaches output a secret shared result.

% Function secret sharing \cite{boyle2015function, boyle2016function} is a paradigm which can be seen as the dual \change{FB: does "dual" have a formal meaning here? TR: not at all, I've added: which can be seen as the dual} of classical data secret sharing, as it provides a different approach to compute similar operations. Indeed, instead of sharing input $x$ into additive random shares $\shared{x}$ and computing a public $f$ onto it to obtain $ \shared{f(x)}$, function secret sharing considers sharing a private function $f$ and applying each function share or \emph{key} $(\shared{f}_0, \shared{f}_1)$ on a public input, namely $f(x) = \shared{f}_0(x) + \shared{f}_1(x) \bmod 2^n$. $f$ is typically a linear operation or a comparison operator. These two approaches are usually equivalent, as illustrated by the following example.

Let us take an example: say Alice and Bob respectively have shares $\shared{y}_0$ and $\shared{y}_1$ of a private input $y$, and they want to compute $\shared{y \le 0}$. 
% While $y\in\Z_{2^n}$, we can associate it to an $n$-bit signed integer in $[-2^{n-1},2^{n-1}-1]$, where the most significant bit (MSB) is a sign bit.
%: positive numbers have their MSB equal to 0 and negative have their MSB equal to 1. 
%They can either use a protocol like SecureNN \cite{wagh2018securenn} which runs directly on the shares $\shared{y}_i$, or 
They first mask their shares using a random mask $\shared{\alpha}$, by
%the operation $f : x \rightarrow (x \ge 0)$ on those shares and produces a shared output $\shared{y \ge 0}$, where each share is held by a party. Or, they can be given shares $\shared{\alpha}$ of a random mask $\alpha$ along with a share $\key_j$ of a function $f_{\alpha} : x \rightarrow (x \ge \alpha)$.
computing $\shared{y}_0 + \shared{\alpha}_0$ and $\shared{y}_1 + \shared{\alpha}_1$, and then reveal these values to reconstruct
$x = y + \alpha$. Next, they apply this public $x$ on their function shares $\shared{f_\alpha}_j$ of $f_{\alpha} : x \rightarrow (x \le \alpha)$, to obtain a shared output $(\shared{f_{\alpha}}_0(x), \shared{f_{\alpha}}_1(x)) = \shared{f_{\alpha}(y + \alpha)} = \shared{(y + \alpha) \le \alpha} = \shared{y \le 0}$. \cite{boyle2015function, boyle2016function} have shown the existence of such function shares for comparison which perfectly hide $y$ and the result. From now on, to be consistent with the existing literature, we will denote the function keys $(\key_0, \key_1) := (\shared{f}_0, \shared{f}_1)$. %\info{Why use two notations $\key_0$ and $\shared{f}_0$? To better stress the parallel with additive value sharing, but keep the existing notation k0 k1, maybe it's clearer now.}

Note that for a perfect comparison, $y+\alpha$ should not wrap around and become negative. Because typically values of $y$ used in practice in machine learning are small compared to the $n$-bit encoding amplitude with typically $n = 32$, the failure rate is less than one comparison in a million, as detailed in Section~\ref{sec:comparison}.

\subsection{2-Party Computation in the Preprocessing Model}
Preprocessing is performed during an offline phase by a trusted third party that builds and distributes the function keys to the 2 parties involved in future computation. This is standard in function secret sharing, and as mentioned by \cite{boyle2019secure}, in the absence of such trusted dealer, the keys can alternatively be generated via an interactive secure protocol that is executed offline, before the inputs are known.
This setup can also be found in other privacy-preserving machine learning frameworks including 
SecureML \cite{mohassel2017secureml}.
This trusted dealer is not active during the online phase, and he is unaware of the computation the 2 parties intend to execute. In particular, as we are in the honest-but-curious model, it is assumed that no party colludes with the dealer.
In practice, such third party would typically be an institution concerned about its reputation, and it could be easy to check that preprocessed material is correct using a \emph{cut-and-choose} technique \cite{zhu2016cut}. For example, the third party produces $n$ keys for private comparison. The 2 parties willing to do the private computation randomly check some of them: they extract from their keys  $s_0$, $s_1$ and also reconstruct $\alpha$ from $\shared{\alpha}_j$, $j \in \zerone$. They can then derive the computations of $\mathsf{KeyGen}$ and verify that the correlated randomness of the keys was correct. They can then use the remaining keys for the private computation.

\subsection{Security Model of the Function Secret Sharing Protocol}
We consider security against \emph{honest-but-curious} adversaries, i.e., parties following the protocol but trying to infer as much information as possible about others' input or function share. This is a standard security model in many SMPC frameworks \cite{bogdanov2008sharemind, ben2016optimizing, riazi2018chameleon, wagh2018securenn} and is aligned with our main use case: parties that would not follow the protocol would face major backlash for their reputation if they got caught. 
The security of our protocols relies on indistinguishability of the function shares, which informally means that the shares received by each party are computationally indistinguishable from random strings. 
More formally, we introduce the following definitions from \cite{boyle2016function}.

\theoremstyle{definition}
\begin{definition}[FSS: Syntax] \label{def:fss_syntax}
A (2-party) \textit{function secret sharing (FSS) scheme} is a pair of algorithms ($\mathsf{KeyGen}$, $\mathsf{Eval}$) with the following syntax:
\begin{itemize}
\item $\mathsf{KeyGen}(1^\lambda, \hat f )$ is a PPT \textit{key generation} algorithm, which on input $1^\lambda$ (security parameter) and $\hat f \in \{0, 1\}^*$, description of a function $f : \Z_{2^n} \rightarrow  \Z_{2^n}$, outputs a pair of keys $(\key_0, \key_1)$.
\item  $\mathsf{Eval}(i, \key_i, x)$ is a polynomial-time \textit{evaluation} algorithm, which on input $i \in \zerone$ (party index), $\key_i$ (the $i$-th function key) and $x \in \Z_{2^n}$, outputs $\shared{f}_i(x) \in \Z_{2^n}$ (the $i$-th share of $f(x)$).
\end{itemize}
\end{definition}

\theoremstyle{definition}
\begin{definition}[FSS: Correctness and Security]
We say that ($\mathsf{KeyGen}$, $\mathsf{Eval}$) as in Definition \ref{def:fss_syntax} is a \textit{FSS scheme for a family of function $\mathcal{F}$} if it satisfies the following requirements:
\begin{itemize}
\item \textbf{Correctness:} For all $f: \Z_{2^n} \rightarrow  \Z_{2^n} \in \mathcal{F}$, $\hat f$ a description of $f$, and $x \in \Z_{2^n}$, if $(\key_0, \key_1) \leftarrow \mathsf{KeyGen}(1^\lambda, \hat f )$ then Pr$[\mathsf{Eval}(0, \key_0, x) + \mathsf{Eval}(1, \key_1, x) = f(x) ] = 1$.
\item  \textbf{Security:} For each $i \in \zerone$, there is a PPT algorithm $\mathsf{Sim}_i$ (simulator), such that for every infinite sequence $(\hat f_j)_{j \in \mathbb{N}}$ of descriptions of functions from $\mathcal{F}$ and polynomial size input sequence $x_j$ for $f_j$, the outputs of the following experiments $\mathsf{Real}$ and $\mathsf{Ideal}$ are computationally indistinguishable:
\begin{itemize}
    \item $\mathsf{Real}_j: (\key_0, \key_1) \leftarrow \mathsf{KeyGen}(1^\lambda, \hat f_j )$ ; Output $\key_i$
    \item $\mathsf{Ideal}_j:$ Output $\mathsf{Sim}_i(1^\lambda)$
\end{itemize}
\end{itemize}
\end{definition}

\cite{boyle2016function} has proved the existence of efficient FSS schemes in particular for equality. Such protocols and the ones that we derive from this work are proved to be secure against semi-honest adversaries, and as mentioned by \cite{boyle2019secure}, they could be extended to guarantee \emph{security with abort} against malicious adversaries using MAC authentication~\cite{damgaard2012multiparty}, which means that the protocol would abort if parties deviated from it.

\subsection{General Security Guarantees and Threats}
The 2-party interaction for private prediction, i.e. when the model is already trained, is an example of \emph{Encrypted Machine Learning as a Service (EMLaaS)}. In this scenario, as stated above, even a malicious model owner could not disclose information about the private inputs or predictions. However, it could use a different model where the weights have been modified to make poor or biased predictions. It is difficult for the data owner to realize that the model owner is misbehaving or using a model whose performance is inferior to what it claims, and this is an issue users also have with standard \emph{Machine Learning as a Service (MLaaS)}. Proving that the computation corresponds to a \emph{certified} given model would require to commit the model and would be costly. On the other side, the information obtained by the data owner about the model (i.e. the prediction on a given input) is the same as in MLaaS. Model inversion techniques \cite{zhang2020secret} can leverage multiple calls to the model to try to build a new model with similar performance. There are not many defenses against this, except limiting access to the model, which is usually the case in MLaaS where data owners are given a quota of requests. Also, attacks like membership inference \cite{shokri2017membership} or reverse-engineering \cite{fredrikson2015model, hitaj2017deep} methods could be used to unveil information about the dataset on which the model was originally trained. Using \emph{differential privacy} \cite{dwork2014algorithmic, abadi2016deep} during the initial training of the model can provide some guarantees \cite{rahman2018membership} against these threats, but it has a trade-off between privacy and utility as differentially private models usually have poorer performance. 

Beyond evaluation, the case of fully-encrypted training can also expose the parties to some threats. The most common one is \emph{data poisoning} and consists of the data owner undermining the training by providing irrelevant data or labels that are wrong or biased \cite{bagdasaryan2020backdoor}. This attack however does not affect privacy. In return, if the model owner gets the final model in plaintext at the end of the training, the privacy of the data owner is at risk because the model owner could use the aforementioned techniques to get information about the training data. Using differential privacy during the private training is important to mitigate this privacy leakage, and should also be applied in a $n$-party training setting.

%\todo[inline]{DP: during training, only the model owner learns the result, so data poisoning from the data owner can only impact correctness of the later evaluations, but not privacy of the model; however the model owner can get information from the data owner: Differential Privacy is important to preserve data-owner privacy.}
%\todo[inline]{TR: Updated}

All these threats must be taken seriously when building production-ready systems. However, they are independent of the function secret sharing protocol and can be addressed separately by combining our work with differential privacy libraries for deep learning.

\section{Function Secret Sharing Primitives}
Our algorithms for private equality and comparison are built on top of the work of \cite{boyle2016function}, so the security assumptions are the same as in this article. We first present an algorithm for equality which is very close to the one of \cite{boyle2016function} but which is used as a basis to build the comparison protocol. We then describe the private comparison protocol, which improves over the work of  \cite{boyle2016function} on Distributed Interval Functions (DIF) by specializing on the operations needed for neural network evaluation or training. In particular, we are able to reduce the function key size from roughly $n(4 \lambda + n)$ to $n(\lambda + 2n)$.

\subsection{Equality Test}

\begin{figure}[tb]
  \centering
  \includegraphics[width=0.9\linewidth]{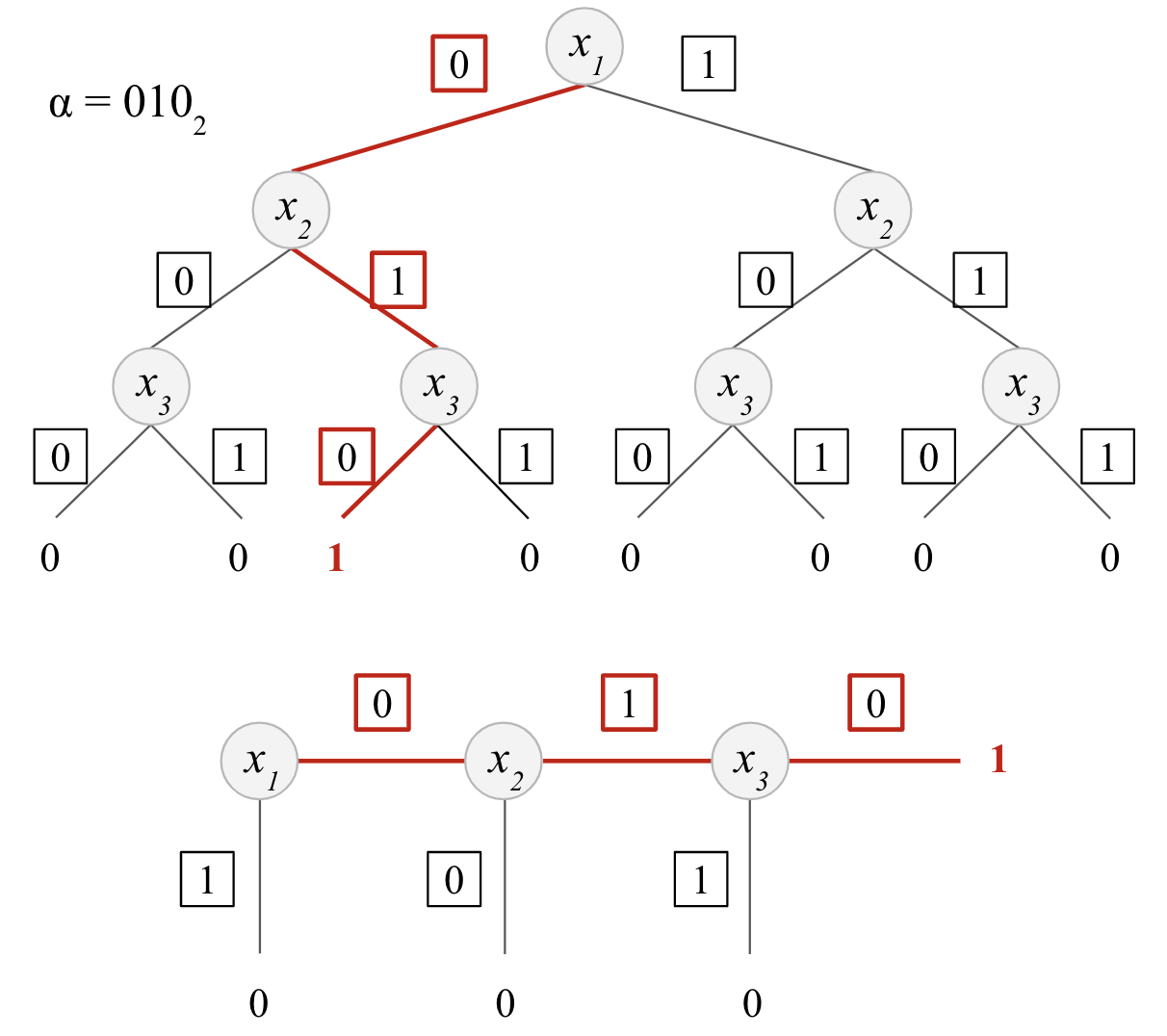}
  \caption{\emph{(Above)} Binary decision tree with the special path for $n=3$. Given an input $x = x[1] \dots x[n]$, at each level $i$, one should take the path labeled by the value in the square equal to the bit value $x[i]$. \emph{(Below)} Flat representation of the tree.}\label{fig:special_path_eq}
\end{figure}

We start by describing private equality as introduced by \cite{boyle2016function}, which is slightly simpler than comparison and gives useful hints about how comparison works.
The equality test consists in comparing a public input $x$ to a private value $\alpha$.
Evaluating the input using the function keys can be viewed as walking a binary tree of depth $n$, where $n$ is the number of bits of the input (typically 32). Among all the possible paths, the path from the root down to $\alpha$ is called the \emph{special path}.
Figure~\ref{fig:special_path_eq} illustrates this tree and provides a compact representation which is used by our protocol, where we do not detail branches for which all leaves are 0.
Evaluation goes as follows: two evaluators are each given a function key which includes a distinct initial random state $(s, t) \in \zerone^{\lambda}\times \zerone$.
%\info{FB: not clear what the initial state is, how states relate to Figure 1}
Each evaluator starts from the root, at each step $i$ goes down one node in the tree and updates his state depending on the bit $x[i]$ using a common \emph{correction word} $\smash{CW^{(i)} \in \zerone^{2(\lambda + 1)}}$ from the function key. At the end of the computation, each evaluator outputs $t$. As long as $x[i] = \alpha[i]$, the evaluators stay on the special path and because the input $x$ is public and common to them, they both follow the same path.
If a bit $x[i] \neq \alpha[i]$ is met, they leave the special path and should output 0 ; else, they stay on it all the way down, which means that $x = \alpha$ and they should output $1$. 

\vspace*{.111cm}
\noindent
\textbf{Intuition.}
The main idea is that while they are on the special path, evaluators should have states $(s_0, t_0)$ and $(s_1, t_1)$ respectively, such that $s_0$ and $s_1$ are i.i.d. and $t_0 \oplus t_1 = 1$. When they leave it, the correction word should act to have $s_0 = s_1$ 
%\todo[inline]{DP: how can they be i.i.d. and $s_0 = s_1$?}
%\todo[inline]{TR: so no it's maybe not i.i.d. but they are indistinguishable from random for the evaluators.}
%\todo[inline]{PT: they are indistinguishable from random \textit{and independent} when they are on the special path. We only ask $s_0 = s_1$ outside the special path so it is not contradictory}
but still indistinguishable from random and $t_0 = t_1$, which ensures $t_0 \oplus t_1 = 0$. To reconstruct the result in plaintext, each evaluator should output its $t_j$ and the result will be given by $t_0 \oplus t_1$. The formal description of the protocol is given below and is composed of two parts: first, in  Algorithm~\ref{algo:key_gen_eq}, the $\mathsf{KeyGen}$ algorithm consists of a preprocessing step to generate the functions keys, and then, in   Algorithm~\ref{algo:eval_eq}, $\mathsf{Eval}$ is run by two evaluators to perform the equality test. It takes as input the private share held by each evaluator and the function key that they have received.
They   use   $\smash{G: \zerone^\lambda \rightarrow \zerone^{2(\lambda + 1)}}$, a pseudorandom generator (PRG), where the output set is $\smash{\zerone^{\lambda + 1} \!\times \! \zerone^{\lambda + 1}}$, and operations modulo $2^n$ implicitly convert back and forth $n$-bit strings into integers.

\begin{algorithm*}[htb] 
    \KwInitialisation{Sample random $\alpha \sampled \Z_{2^n}$ \\
    Sample random $s^{(1)}_j \sampled \zerone^\lambda$ and set $t_{j}^{(1)} \leftarrow j$, for $j = 0,1$}
    \For{$i = 1..n$}{
        $(s_j^\mathsf{L} \sep t_j^\mathsf{L}, s_j^\mathsf{R} \sep t_j^\mathsf{R}) \leftarrow G(s^{(i)}_j)\in \zerone^{\lambda+1} \times \zerone^{\lambda+1}$,  for $j = 0,1$ \\
        \leIf{$\alpha[i]$}{
            $cw^{(i)} \leftarrow ( 0^\lambda \sep 0, s_0^\mathsf{L} \oplus s_1^\mathsf{L} \sep 1)$ \\}{
            $cw^{(i)} \leftarrow ( s_0^\mathsf{R} \oplus s_1^\mathsf{R} \sep 1, 0^\lambda \sep 0)$} 
        $\CWi \leftarrow cw^{(i)} \oplus G(s^{(i)}_0) \oplus G(s^{(i)}_1) \in \zerone^{\lambda+1} \times \zerone^{\lambda+1}$ \\
        $state_j \leftarrow G(s^{(i)}_j) \oplus (t_j^{(i)} \cdot \CWi) = (state_{j,0}, state_{j,1})$, for $j = 0,1$ \\
        Parse $s_j^{(i+1)} \sep t_j^{(i+1)} = state_{j,\alpha[i]}\in \zerone^{\lambda+1}$, for $j = 0,1$
    }
    $CW^{(n+1)} \leftarrow (-1)^{t_1^{(n+1)}} \cdot \left(1 - s_0^{(n+1)} + s_1^{(n+1)}\right) \bmod 2^n$ \\
    \KwRet $\key_j \leftarrow \shared{\alpha}_j \sep s_j^{(1)} \sep CW^{(1)} \sep \cdots \sep CW^{(n+1)}$, for $j = 0,1$
\caption{$\mathsf{KeyGen}$: function key generation for equality (from \cite{boyle2016function}) \label{algo:key_gen_eq}} 
\end{algorithm*}
\begin{algorithm*}[htb] 
    \KwIn{$(j, \key_j, \shared{y}_\bool)$ where $\bool \in \zerone$ refers to the evaluator id}
    Parse $\key_j$ as $\shared{\alpha}_j \sep s^{(1)} \sep CW^{(1)} \sep \cdots \sep CW^{(n+1)}$ \\
    Publish $\shared{\alpha}_j + \shared{y}_j \bmod 2^n$ and get revealed $x = \alpha + y \bmod 2^n$ \\ 
    Let $t^{(1)} \leftarrow j$ \\
    \For{$i = 1..n$}{
        $state \leftarrow G(s^{(i)}) \oplus (t^{(i)} \cdot \CWi) = (state_0, state_1)$ \\
        Parse $s^{(i+1)} \sep t^{(i+1)} = state_{x[i]}$
    }
    \KwRet $\shared{T}_j \leftarrow (-1)^j \cdot \left(t^{(n+1)} \cdot CW^{(n+1)} + s^{(n+1)}\right) \bmod 2^n$ 
\caption{$\mathsf{Eval}$: evaluation of the function key for the equality test $y = 0$ (from \cite{boyle2016function}) \label{algo:eval_eq}}
\end{algorithm*}

\vspace*{.111cm}
\noindent
\textbf{Correctness.}
Intuitively, the correction words $CW^{(i)}$ are built from the expected state of each evaluator on the special path, i.e., the state that each should have at each node $i$ if it is on the special path given some initial state. During evaluation, a correction word is applied by an evaluator only when it has $t = 1$. Hence, on the special path, the correction is applied only by one evaluator at each bit. If at step $i$, the evaluator stays on the special path, the correction word compensates the current states of both evaluators by xor-ing them with themselves and re-introduces a pseudorandom value $s$ (either $\smash{s_0^\mathsf{R} \oplus s_1^\mathsf{R}}$ or $\smash{s_0^\mathsf{L} \oplus s_1^\mathsf{L}}$), which means the xor of their states is now $(s, 1)$ but those states are still indistinguishable from random.

%randomized by the other halves of the pseudorandom values.
On the other hand, if $x[i] \neq \alpha[i]$, the new state takes the other half of the correction word, so that the xor of the two evaluators states is (0, 0). From there, they have the same states and both have either $t = 0$ or $t = 1$. They will continue to apply the same corrections at each step and their states will remain the same, meaning that $t_0 \oplus t_1 = 0$. A final computation is performed to obtain a shared $\shared{T}$ modulo $2^n$ of the result bit $t = t_0 \oplus t_1\in\zerone$.
% \todo{FB: do we need details of the algorithms  in the main text?}

\vspace*{.111cm}
\noindent
\textbf{Security.}
From the privacy point of view, when the seed $s$ is random, $G(s)$ is indistinguishable from random (this is a pseudorandom bit-string). Each half is used either in the $cw$ or in the next state, but not both. 
Therefore, the correction words $CW^{(i)}$ do not contain information about the expected states and for $j=0,1$, the output $\key_j$ is independently uniformly distributed with respect to $\alpha$ and $\smash{s^{\small{(1)}}_{1-j}}$, in a computational way.
%Therefore, the correction words $CW^{(i)}$ do not contain information about the expected states: from the initial $n+2\lambda$ random bits ($\alpha \sampled \Z_{2^n}$ and $(s^{(1)}_0, s^{(1)}_1) \sampled \zerone^\lambda \times \zerone^\lambda$), the output $\key_j$ (a $((4+2\lambda) n + \lambda)$-bit string) and $\alpha$ are i.i.d, for $j=0,1$, in a computational way.
As a consequence, at the end of the evaluation, for $j=0,1$, $\shared{T}_j$ also follows a distribution independent of $\alpha$. Until the shared values are reconstructed, even a malicious adversary cannot learn anything about $\alpha$ nor the inputs of the other player.

\vspace*{.111cm}
\noindent
\textbf{Implementation.}
Function keys should be computed by a third party dealer and sent to the evaluators in advance, which requires one extra communication of the size of the keys. We use the trick of \cite{boyle2016function} to reduce the size of each correction word in the keys, from $2(1 + \lambda)$ to $(2 + \lambda)$ by reusing the pseudo-random $\lambda$-bit string dedicated to the state used when leaving the special path for the state used for staying onto it, since for the latter state the only constraint is the pseudo-randomness of the bitstring. Regarding the PRG, we use a Matyas-Meyer-Oseas one-way compression function with an AES block cipher, as in \cite{keller2016mascot} or \cite{wang2017splinter}. We concatenate several fixed key block ciphers to achieve the desired output length: $G(x) = E_{k_1}(x) \oplus x \sep E_{k_2}(x) \oplus x \sep \dots$. Using AES helps us to benefit from hardware acceleration: we used the \verb|aesni| Rust library for CPU execution and the \verb|csprng| library of PyTorch for GPU. More details about implementation can be found in Appendix \ref{app:prg}.

\subsection{Comparison}
\label{sec:comparison}

Our main contribution to the function secret sharing scheme is for the comparison function, which is intensively used in neural network to build non-polynomial activation functions like ReLU: we build on the idea of the equality test to provide a synthetic and efficient protocol whose structure is very close to the previous one, and improves upon the former DIF scheme of \cite{boyle2016function} by significantly reducing the key size.

\begin{figure}[htb!]
  \centering
  \includegraphics[width=0.9\linewidth]{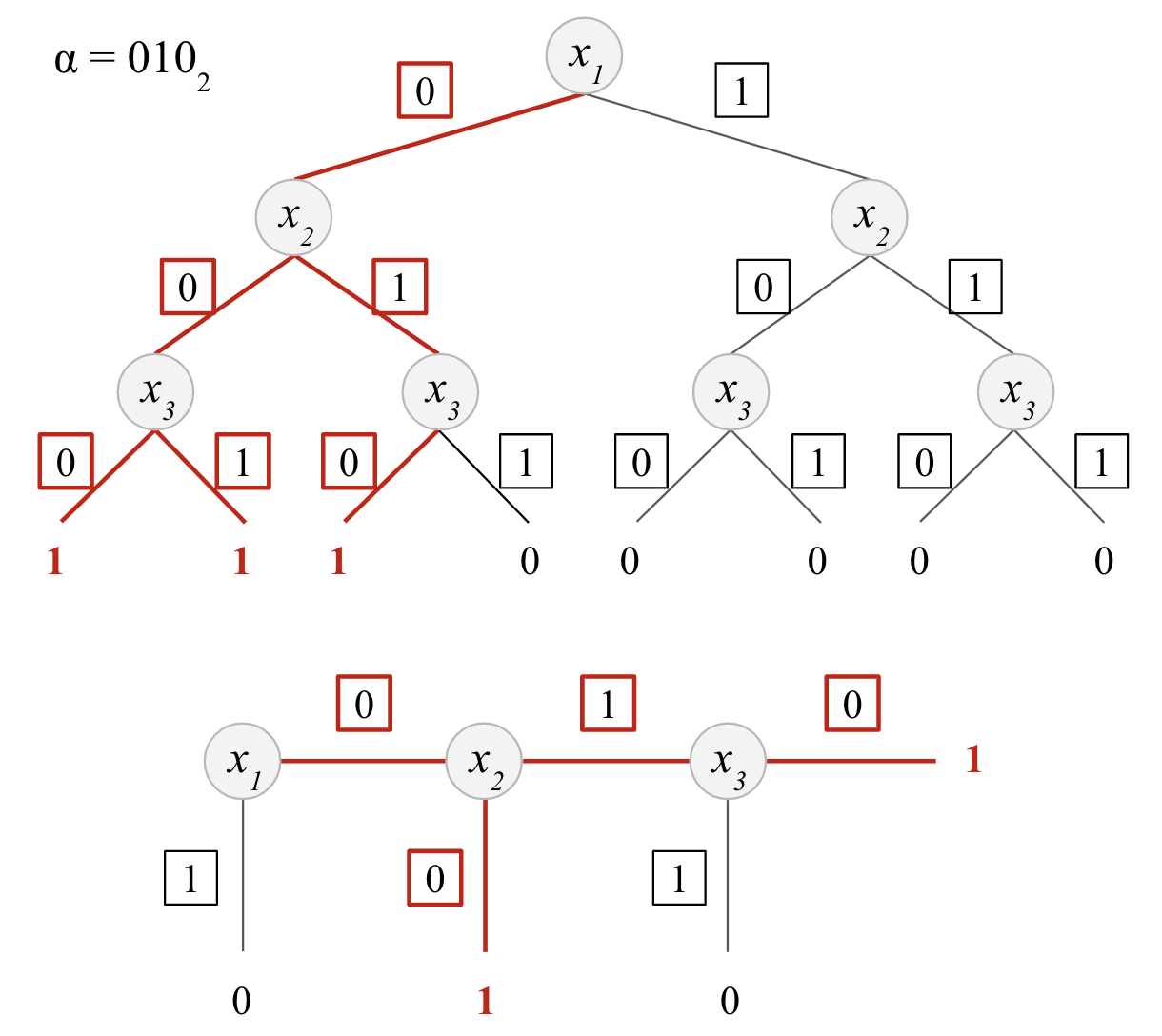}
  \caption{\emph{(Above)} Binary decision tree with all the paths corresponding to $x \le \alpha$ for n=3. \emph{(Below)} Flat representation of the tree.}\label{fig:special_path_comp}
\end{figure}

\subsubsection{Intuition}

Instead of seeing the special path as a simple path, we can see it as a frontier for the zone in the tree where $x \le \alpha$. To evaluate $x \le \alpha$, we could evaluate all the paths on the left of the special path and then sum up the results, but this is highly inefficient as it requires exponentially many evaluations. The key idea here is to evaluate all these paths at the same time, noting that each time one leaves the special path, it either falls on the left side (i.e., $x < \alpha)$ or on the right side (i.e., $x > \alpha$). Hence, we only need to add an extra step at each node of the evaluation, where depending on the bit value $x[i]$, we output a leaf label which is 1 only if $x[i] < \alpha[i]$ and all previous bits are identical. % (because getting on this path means that we were on the special path until the node $i - 1$).
%Because this happens privately, evaluators don't know when the decision is made and need to compute all the $n$ nodes. However, o
Only one label between the final label (which  corresponds to $x = \alpha$) and the leaf labels can be equal to one, because only a single path can be taken. Therefore, evaluators will return the sum of all the labels to get the final output.

\subsubsection{Correctness}

\textbf{Correctness of the comparison protocol.} Consider $(k_0, k_1)$ generated by $\mathsf{KeyGen}$ (Algorithm~\ref{algo:key_gen_comp}) with a random offset $\alpha \in \mathbb{Z}_{2^n}$. Consider a public input $x \in \mathbb{Z}_{2^n}$. Let us show that $\mathsf{Eval}(0, k_0, x) + \mathsf{Eval}(1, k_1, x) = (x \le \alpha) \mod 2^n$, where $(x \le \alpha) \in \{0,1\}$. 
We add a subscript $0$ or $1$ to the variables of Algorithm~\ref{algo:eval_comp} to identify the party to which they belong. 

Consider $i \in [1,n]$ such that the evaluators remained on the special path until $i$ (\textit{i.e.} $\forall j < i, x[j] = \alpha[j]$). In particular, $G(s^{(i)}_0) \xor G(t^{(i)}_0 \cdot CW^{(i)}) \xor (s^{(i)}_1) \xor (t^{(i)}_1 \cdot CW^{(i)}) = cw^{(i)}$.
Let us study the 4 possible cases and show that 1) $(out_{i,0} + out_{i,1} \mod 2^n) \in \{0,1\}$; 2) $out_{i,0} + out_{i,1} = 1 \mod 2^n \iff x[i] < \alpha[i]$; and 3) the evaluators stay on the special path if and only if $x[i] = \alpha[i]$.

\begin{itemize}
    \item If $x[i] = 0$, we keep the left part of $state'$ at line 4.
    
    \begin{itemize}
    \item  If $\alpha[i] = 1$, we have $\tau^{(i+1)}_0 \xor \tau^{(i+1)}_1 = 1$. 
Thanks to line~\ref{line:comp_keygen_10} of $\mathsf{KeyGen}$, we have $out_{i,0} + out_{i,1} = \left(\tau_0^{(i+1)} - \tau_1^{(i+1)}\right)CW^{(i)}_{leaf} +  (\sigma^{(i+1)}_0 - \sigma^{(i+1)}_1) = (1-2\tau_1^{(i+1)}) (-1)^{\tau_1^{(i+1)}} (\sigma^{(i+1)}_1 - \sigma^{(i+1)}_0 + 1) +  (\sigma^{(i+1)}_0 - \sigma^{(i+1)}_1) = 1 \mod 2^n$. 
We also have $t^{(i+1)}_0 \xor t^{(i+1)}_1 = 0$ and $s^{(i+1)}_0 \xor s^{(i+1)}_1 = 0$, so the evaluators leave the special path.

% $CW^{(i)_{leaf}} = (-1)^{0} \cdot \left( 0 - 0 +  1 \right)$. 
% Hence, $out_{i,0} + out_{i,1} = \tau_0 \oplus \tau_1 \mod 2^n$.
% (-1)^{\tau_1^{(i+1)}}\left(\sigma^{(i+1)}_1 - \sigma^{(i+1)}_0 + \alpha[i]\right)

 \item  If $\alpha[i] = 0$, we use line 5 of $\mathsf{KeyGen}$ to generate $cw^{(i)}$, so $\sigma^{(i+1)}_0 \xor \sigma^{(i+1)}_1 =0$ and $\tau^{(i+1)}_0 \xor \tau^{(i+1)}_1 = 0$. 
Hence,  $out_{i,0} + out_{i,1} = \left(\tau_0^{(i+1)} - \tau_1^{(i+1)}\right)CW^{(i)}_{leaf} +  (\sigma^{(i+1)}_0 - \sigma^{(i+1)}_1) = 0 \mod 2^n$.
We also have $t^{(i+1)}_0 \xor t^{(i+1)}_1 = 1$ and $s^{(i+1)}_0, s^{(i+1)}_1$ stay on the special path.

\end{itemize}

\item If $x[i] = 1$, we keep the right part of $state'$ at line 4.

\begin{itemize}
    \item If $\alpha[i] = 1$, we use line 4 of $\mathsf{KeyGen}$  to generate $cw^{(i)}$, so $\sigma^{(i+1)}_0 \xor \sigma^{(i+1)}_1 =0$ and $\tau^{(i+1)}_0 \xor \tau^{(i+1)}_1 = 0$.
Hence, similarly as the case where $(x[i], \alpha[i]) = (0,0)$, we have $out_{i,0} + out_{i,1} = 0 \mod 2^n$ and the evaluators stay on the special path.

    \item If $\alpha[i] = 0$, we use line 5 of $\mathsf{KeyGen}$  and get $cw^{(i)} = \big( (s_0^\mathsf{R} \oplus s_1^\mathsf{R} \sep 1, 0^\lambda \sep 0), 
        (0^\lambda \sep 0, \sigma_0^\mathsf{L} \oplus \sigma_1^\mathsf{L} \sep 1) \big)$. We keep the right part of $state'$ at line~\ref{line:comp_keygen_9} of $\mathsf{KeyGen}$ for $CW^{(i)}_{leaf}$.
        We use the same right part at line 5 of $\mathsf{Eval}$, so we have $\tau_0^{(i+1)} \xor \tau_1^{(i+1)} = 1$.
        Finally, $out_{i,0} + out_{i,1} = \left(\tau_0^{(i+1)} - \tau_1^{(i+1)}\right)CW^{(i)}_{leaf} +  (\sigma^{(i+1)}_0 - \sigma^{(i+1)}_1) = (1-2\tau_1^{(i+1)}) (-1)^{\tau_1^{(i+1)}} (\sigma^{(i+1)}_1 - \sigma^{(i+1)}_0 + 0) +  (\sigma^{(i+1)}_0 - \sigma^{(i+1)}_1) = 0 \mod 2^n$.
        We also have  $t^{(i+1)}_0 \xor t^{(i+1)}_1 = 0$ and the evaluators leave the special path.
\end{itemize}
\end{itemize}

If the evaluators leave the special path at step $i$, their bistrings remain equal until the end of the evaluation: $\forall j \in [i, n+1], s^{(j)}_0 =  s^{(j)}_1$  and $\sigma^{(j)}_0 =  \sigma^{(j)}_1$, so $\forall j \in [i, n+1], out_{j,0} + out_{j,1} = 0 \mod 2^n$.

Finally, if the evaluators never leave the special path (\textit{i.e.} $x = \alpha$), we have $\forall j \in [1, n], out_{j,0} + out_{j,1} = 0 \mod 2^n$, and $out_{n+1,0} + out_{n+1,1} = 1$. Indeed, step $n+1$ is identical to the equality case (Algorithm \ref{algo:eval_eq}). 

In the end, the sum $\shared{T}_j$ of the $out_i$'s is a share of 1 either if $out_{n+1}$ was a share of 1 (i.e. $x = \alpha$) or if one of the other $out_i$ was a share of 1, which is possible only if $\alpha[i] = 1$ and $x[i] < \alpha[i]$  (i.e. $x < \alpha$). Otherwise (i.e.  $x > \alpha$), $\shared{T_j}$ is a share of 0.

% \todo[inline]{1) Is it formal enough? 2) Should I use the "proof" environment and state this as a theorem/claim? TR: for me this looks good! But I'm not a formalism expert}

\begin{algorithm*}[htb]
    \KwInitialisation{Sample random $\alpha \sampled \Z_{2^n}$ \\ 
    Sample random $s^{(1)}_j \sampled \zerone^\lambda$ and set $t_{j}^{(1)} \leftarrow j$, for $j = 0,1$} 
    \For{$i = 1..n$}{
        \For{$j = 0,1$}{
            $( (s_j^\mathsf{L} \sep t_j^\mathsf{L}, s_j^\mathsf{R} \sep t_j^\mathsf{R}),
            (\sigma_j^\mathsf{L} \sep \tau_j^\mathsf{L}, \sigma_j^\mathsf{R} \sep \tau_j^\mathsf{R}) ) \leftarrow G(s^{(i)}_j) \in \zerone^{\lambda+1} \times \zerone^{\lambda+1} \times \zerone^{n+1} \times \zerone^{n+1}$ \\
        }
        \leIf{$\alpha[i]$}{
            $cw^{(i)} \leftarrow \big( (0^\lambda \sep 0,  s_0^\mathsf{L} \oplus s_1^\mathsf{L} \sep 1),
                (\sigma_0^\mathsf{R} \oplus \sigma_1^\mathsf{R} \sep 1, 0^\lambda \sep 0) \big)$ \\}{
            $cw^{(i)} \leftarrow \big( (s_0^\mathsf{R} \oplus s_1^\mathsf{R} \sep 1, 0^\lambda \sep 0), 
        (0^\lambda \sep 0, \sigma_0^\mathsf{L} \oplus \sigma_1^\mathsf{L} \sep 1) \big)$}
        $\CWi \leftarrow cw^{(i)} \oplus G(s^{(i)}_0) \oplus G(s^{(i)}_1)$ \\
        \For{$j = 0,1$}{
            $state_j \leftarrow G(s^{(i)}_j) \oplus (t_j^{(i)} \cdot \CWi) = ((state_{j,0}, state_{j,1}),(state'_{j,0}, state'_{j,1}))$ \\
            Parse $s_j^{(i+1)} \sep t_j^{(i+1)} = state_{j, \alpha[i]}$
            and $\sigma_j^{(i+1)} \sep \tau_j^{(i+1)} = state'_{j, 1-\alpha[i]}$ \label{line:comp_keygen_9}
        }
        $\CWi_{leaf} \leftarrow (-1)^{\tau_1^{(i+1)}} \cdot \left(\alpha[i] - \sigma_0^{(i+1)} + \sigma_1^{(i+1)}  \right) \bmod 2^n$ \label{line:comp_keygen_10}
    }
    $CW^{(n+1)}_{leaf} \leftarrow (-1)^{t_1^{(n+1)}} \cdot \left( 1 - s_0^{(n+1)} + s_1^{(n+1)} \right) \bmod 2^n$ \\
    \KwRet $\kb \leftarrow \shared{\alpha}_j \sep s_j^{(1)} \sep (CW^{(i)})_{i=1..n} \sep (CW_{leaf}^{(i)})_{i=1..n+1}$, for $j = 0,1$
\caption{$\mathsf{KeyGen}$: function key generation for comparison $x \le \alpha$ (new)}
\label{algo:key_gen_comp}
\end{algorithm*}

\begin{algorithm*}[htb]
    \KwIn{$(j, \key_j, x)$ where $\bool \in \zerone$ refers to the evaluator id}
    Parse $\key_j$ as $\shared{\alpha}_j \sep s^{(1)} \sep (CW^{(i)})_{i=1..n} \sep (CW_{leaf}^{(i)})_{i=1..n+1}$ \\
    Let $t^{(1)} \leftarrow j$ \\
    \For{$i = 1..n$}{
        $state \leftarrow G(s^{(i)}) \oplus (t^{(i)} \cdot \CWi) = ((state_0, state_1), (state'_0, state'_1))$ \\
        Parse $s^{{(i+1)}} \sep t^{{(i+1)}} = state_{x[i]}$ and $ \sigma^{(i+1)} \sep \tau^{(i+1)} = state'_{x[i]}$ \\
        $out_i \leftarrow (-1)^\bool \cdot \left( \tau^{(i+1)} \cdot CW_{leaf}^{(i)} + \sigma^{(i+1)} \right) \bmod 2^n$ \\
    }
    $out_{n+1} \leftarrow (-1)^j \cdot \left( t^{(n+1)} \cdot CW_{leaf}^{(n+1)} + s^{(n+1)} \right) \bmod 2^n$ \\
    \KwRet $\shared{T}_j \leftarrow \sum_i out_i \bmod 2^n$
\caption{$\mathsf{Eval}$: evaluation of the function key for comparison $x \le \alpha$ (new)}
\label{algo:eval_comp}
\end{algorithm*}

\begin{algorithm*}[htb]
    \KwIn{$(j, \key_j, \shared{y}_\bool)$ where $\bool \in \zerone$ refers to the evaluator id}
    Parse the first $n$ bits of $\key_j$ as $\shared{\alpha}_j $ \\
    Publish $\shared{\alpha}_j + \shared{y}_j$ and get revealed $x = \alpha + y \bmod 2^n$ \\
    \KwRet $\shared{T}_j \leftarrow \mathsf{Eval}(j, k_j, x)$
\caption{$\mathsf{Sign}$: protocol for $\mathsf{sign}(\shared{y})$}
\label{algo:protocol_comp}
\end{algorithm*}

\textbf{Failure rate of the sign protocol.}\label{sec:failure_rate}
Algorithm~\ref{algo:protocol_comp} details how we build a sign protocol thanks to our comparison primitive (Algorithm~\ref{algo:eval_comp}), following the secret sharing workflow introduced in Section~\ref{sec:background:fss}.
Our sign protocol can fail if $y + \alpha$ wraps around and becomes negative. We cannot act on $\alpha$ because it must be completely random to act as a perfect mask and to make sure the revealed $x = y + \alpha \bmod 2^n$ does not leak any information about $y$, but the smaller $y$ is, the lower the error probability will be. \cite{boyle2019secure} suggests a method which uses 2 invocations of the protocol to guarantee perfect correctness but because it incurs an important runtime overhead, we rather show that the failure rate of our comparison protocol is very small and is reasonable in contexts that tolerate a few mistakes, as in machine learning. 
% \noindent\textbf{Failure probability.} 
Consider $y\in [-2^{n-1}, 2^{n-1}-1]$, a pair of comparison keys $(k_0, k_1)$, and note $\widehat{\mathsf{Sign}}(y) := \mathsf{Sign}(0, k_0, \shared{y}_0) +\mathsf{Sign}(1, k_1, \shared{y}_1)  \mod 2^n$ the reconstructed result of the sign protocol.
%   $\mathsf{Sign}(y) \in \{0,1\} $.
We have $\Pr\left[ \widehat{\mathsf{Sign}}(y)  \neq \mathbb{1}[y \le 0]\right] = |y|/2^n \le Y/2^n$ where $Y$ is the maximum amplitude for $|y|$.

% Indeed, consider $\alpha \in [0,2^{n}-1]$ as defined in Algorithm~\ref{algo:protocol_comp}.
% The protocol calls the comparison primitive to compute $\mathbb{1}[y + \alpha \mod 2^n \le \alpha]$. If $y \le 0$, $y + \alpha \mod 2^n > \alpha$ for $\alpha \in [0, |y|-1]$, while if $y > 0$ it wraps around when $\alpha \in [2^{n}-y, 2^{n}-1]$.
% In the end, the sign protocol fails with probability $\Pr_\alpha[\mathbb{1}[y + \alpha \mod 2^n \le \alpha] \neq \mathbb{1}[y \le 0]] = |y|/2^n $, since $\alpha$ is sampled uniformly at random in Algorithm~\ref{algo:key_gen_comp}.

% \noindent\textbf{Empirical validation.}
We quantify this failure rate on real world examples, namely on Network-2 and on the $64\!\times\!64$ Tiny Imagenet version of VGG16, with a fixed precision of 3 decimals, and find respective failure rates of 1 in 4 millions comparisons and 1 in 100 millions comparisons, which is low compared to the number of comparisons needed for an evaluation, respectively $\sim10$K and $\sim1$M. In practice, such error rates do not affect the model accuracy, as Table \ref{table:private_accuracy} shows. 
% Interestingly, \cite{boylefunction2020} proposes a way to combine the 2 invocations of \cite{boyle2019secure} in a single evaluation while keeping the key size roughly the same. This idea could be useful in scenarios where the error rate wouldn't be acceptable, like when comparing huge numbers.

\subsubsection{Security}

The formal proof of security is provided in Appendix \ref{appendix:security}.

\subsubsection{Implementation and Communication Complexity}
In all these computations modulo $2^n$, the bitstrings $\smash{s_j^{(i)}}$ and $\smash{\sigma_j^{(i)}}$ are respectively in $\zerone^\lambda$ and $\zerone^n$, where we have typically $\lambda = 128$ and $n = 32$. 
The PRG used here is $G: \zerone^\lambda \rightarrow \zerone^{2(\lambda + 1) + 2(n+1)}$ where the output is seen as a pair of pairs of elements in $\left(\zerone^{\lambda+1} \times \zerone^{\lambda+1}\right) \times \left(\zerone^{n+1} \times \zerone^{n+1}\right)$. For the right-hand part, we only need $n$ bits instead of $\lambda$ bits since the $\sigma^{(i)}$ deriving from the PRG are not used for anything other than masking the $n$-bit output. This allows us to use fewer AES block ciphers in our PRG implementation and hence to achieve faster computation.
In addition, because our comparison protocol works very similarly to the equality protocol, we can reuse the trick that consists of reusing randomness of the state corresponding of leaving the special area for the state corresponding of staying into it, as it does not compromise the fact that this state only needs to be pseudo-random. Thanks to this, we almost divide by 2 the size of the $\CWi$ from $2(\lambda + 1) + 2(n + 1)$ to $\lambda + 2 + n + 2$. Compared to the previous Distributed Interval Function (DIF) protocol of \cite{boyle2016function}, our algorithm is not only much simpler as it does not require inspecting binary trees and searching for paths, but it also reduces significantly the key size from roughly $n(4 \lambda + n)$ to $n(\lambda + 2n + 4) + \lambda + 2n$ bits. This allows for faster transmission of keys over the network to the parties doing the evaluation. 
% We note that \cite{boylefunction2020} achieves similar key size for comparison, with $n(\lambda + n + 2) + \lambda + 2n$ bits.

\section{Application to Deep Learning}\label{section:fss_to_ml}
We now apply these primitives to a private deep learning setup in which a model owner interacts with a data owner. The data and the model parameters are sensitive and are secret shared to be kept private. The shape of the input and the architecture of the model are however public, which is a standard assumption in secure deep learning \cite{liu2017oblivious,mohassel2017secureml}. %, as privacy only concerns the parameters of the model once it is trained for a specific task.

% \subsection{Additive sharing workflow}
\subsection{Additive Sharing Workflow with Preprocessing}

% Figure \ref{fig:additive_workflow} illustrates this with a basic workflow
%\begin{figure}[htb!]
%  \centering
%  \includegraphics[width=1\linewidth]{figures/additive_workflow.png}
%  \caption{Additive sharing workflow for the functionality $f: x, z \mapsto (x \le 0)\cdot z$}\label{fig:additive_workflow}
%\end{figure}

%Despite the paradigm induced by function secret sharing, our workflow relies heavily on additive sharing. Indeed, to ensure modularity, all operations follow the same structure:
All our operations are modular and follow this additive sharing workflow: inputs are provided secret shared and are masked with random values before being revealed. This disclosed value is then consumed with preprocessed function keys to produce a secret shared output. Each operation is independent of all surrounding operations, which is known as \emph{circuit-independent preprocessing} \cite{boyle2019secure} and implies that key generation can be fully outsourced without having to know the model architecture. This results in a fast runtime execution with a very efficient online communication, with a single round of communication and a message size equal to the input size for comparison. 

Additionally, values need to be converted from float to fixed point precision before being secret shared. The fixed point representation allows one to store decimal values with some approximation using $n$-bits integers. For example, when using a fixed precision of 3, a decimal value $x$ is stored as $\left \lfloor x \cdot 10^3  \right \rfloor$ in $\Z_{2^n}$. Fixed precision is used to simplify operations like addition because the inputs can be summed up directly in $\Z_{2^n}$.

% Preprocessing is very convenient for all pointwise operations like comparison: one can produce in advance $M$ sub-primitives and each time it needs to evaluate ReLU on a vector of $m$ values it can pick $m$ preprocessed sub-primitives from the global stack. For other operations like matrix multiplication which we detail below, shapes should be known in advance, which can easily be done by evaluating the model on some dummy input. Therefore, preprocessing is also circuit independent for vectorized operations. %neural network layers such as fully connected layers or activation functions. % Not true for convolution

\subsection{Common Machine Learning Operations}\label{sec:common_ml_op}

\textbf{ReLU}
activation function is supported as a direct application of our comparison protocol, which we combine with a point wise multiplication. As mentioned in Section \ref{section:background}, this construction is not exact and is associated with an error rate which is below 1 in a million for typical ML computations. The comparison made in Table \ref{table:private_accuracy} between fixed point and private evaluation of pre-trained models shows that this error rate does not affect model accuracy.
% For more complex activation functions such as softmax, we could use polynomial approximations methods, which achieve acceptable accuracy despite involving a higher number of rounds \cite{ pan1991improved, higham2002accuracy, hannun2019privacy}.

\vspace*{.111cm}

\noindent
\textbf{Matrix Multiplication (MatMul)}, as mentioned by \cite{boyle2019secure}, fits in this additive sharing workflow. We use Beaver triples \cite{beaver1991efficient} to compute $\shared z = \shared{x \cdot y}$ from $\shared x$, $\shared y$ and using a triple $(\shared a, \shared b, \shared c := \shared{a \cdot b})$, where all values are secret shared in $\Z_{2^n}$. The mask is here $\shared{(-a, -b)}$ and is used to reveal $\smash{(\delta, \epsilon) := (x - a, y - b)}$. The functional keys are the shares of $(\shared a, \shared b, \shared c)$ and are used to compute $\smash{\delta \cdot \shared b  + \epsilon \cdot \shared a + \delta \cdot \epsilon + \shared c = \shared z}$. Matrix multiplication is identical but uses matrix Beaver triples \cite{mohassel2017secureml}.

\vspace*{.111cm}
\noindent
\textbf{Convolution}
 can also be computed using Beaver triples. Using the previous notations, we can now consider $y$ to be the convolution kernel, and the operation $\cdot$ now stands for the convolution operator. We use this method for the CPU and GPU implementations, which enables us to use the PyTorch Conv2d function to compute the $\cdot$ operation. Note that convolution can also be computed as a matrix multiplication using an unrolling technique as described in \cite{chellapilla2006high}, but it incurs an overhead in terms of communication because the unrolled matrix is bigger than the original one when the stride is smaller than the kernel size. More details about unrolling can be found in Appendix \ref{app:unrolling} with Figure \ref{fig:unrolling}.

\vspace*{.111cm}
\noindent
\textbf{Argmax} is used to determine the predicted label for classification tasks (\textit{i.e.} compute the index of the highest value of the last layer). Algorithm~\ref{algo:argmax} shows how to compute this operator in a constant number of rounds using pairwise comparisons, in a fashion similar to \cite{hannun2019privacy}. This algorithm outputs the indices in the one-hot format, meaning that the output vector is of a similar shape to the input, and contains $1$ where the maximum was found and $0$ elsewhere. This protocol does not guarantee one-hot output: if the last layer outputs two identical maximum values, both will be retrieved. This sounds acceptable for machine learning evaluation as it informs that the model cannot choose between two classes. For training, the output signal only needs to be normalized. Probabilistic techniques are available to break ties, which only require an additional comparison. 

% The main idea here is, given a vector $\smash{(x_0, \dots, x_{m-1})}$, to compute the matrix 
% $\smash{M \in \mathbb{R}^{m-1 \times m}}$ where each row $i$ is defined as such:
% $\smash{M_i = (x_{i + 1 \mod m}, \dots,  x_{i + m + 1  \mod m})}$. 
% Then, all the elements of column $j$ (i.e. the $x_i$ with $i \ne j$) are compared to $x_j$, which requires $m(m-1)$ parallel comparisons. A column $j$ where all elements are lower than $x_j$ indicates that $j$ is a valid result for the argmax. Identifying such valid $j$ is done by summing over all values in the column $j$ resulting from the previous comparison and checking that the sum equals $m - 1$. 

In our algorithm, the first loop (line~\ref{line:first_loop}) requires $m(m-1)$ parallel comparisons, and the second loop (line~\ref{line:second_loop}) requires $m$ equality checks.
Hence, the argmax uses 2 rounds of communication and sends $O(m^2)$ values over the network. This is reasonable for a neural network where the number of outputs $m$ is about $100$ or less.

% \begin{algorithm*}[!htb]
% Weird LateX error (but compiles)
\begin{algorithm}[t]
    \KwIn{$\shared{x_1}, \dots, \shared{x_m}$}
    \KwOut{$\arg\max_{i \in [1,m]} x_i$}
    \For{$j \in [1,m]$}{
        $\shared{s_j} \leftarrow \sum_{i \neq j} \shared{x_i - x_j \le 0}$\label{line:first_loop}
    }
    \For{$j \in [1,m]$}{
        $\shared{\delta_j} \leftarrow \shared{s_j = m - 1}$\label{line:second_loop}
    }
    % $\operatorname{Reveal} \delta_1, \dots, \delta_m \in \{0,1\}$\\
    \KwRet $\shared{\delta_1}, \dots, \shared{\delta_m}$
\caption{$\mathsf{Argmax}$ functionality using FSS}
\label{algo:argmax}
\end{algorithm}

\vspace*{.111cm}
\noindent
\textbf{MaxPool}
can be implemented by combining the ideas of the unrolling-based convolution and the argmax: the matrix is first unrolled like in Figure \ref{fig:unrolling} and the argmax of each row is then computed using parallel pairwise comparisons. This argmax is then multiplied with the row to get the maximum value, and the matrix is rolled back. These steps are illustrated in Figure \ref{fig:maxpool} in Appendix \ref{app:maxpool} and is formally described in Algorithm \ref{algo:maxpool}. It requires 3 rounds of communication, but we also provide an optimization when the kernel size $k$ equals 2, which reduces the computation complexity by a factor $4\times$ but uses an additional round of communication, and is very useful for some deep models such as VGG16.

\begin{algorithm}
    \KwIn{$\shared{\mathbf{X}} = (\shared{x_{i, j}})_{i, j =1 \dots m}$}
    \KwOut{$\shared{\mathsf{MaxPool}(\mathbf{X}, k)}$}
    Set $n = \lfloor(m-k)/s+1\rfloor$ \\
    Define $\shared{\mathbf{X}^{\mathsf{unrolled}}}$ of shape $n^2 \times k^2$\\
    Define $\shared{\vec y}$ of size $n^2$\\
    \For{$i,j \in [0, s, 2s, \dots, m-k]$}{
        $\shared{x^{\mathsf{unrolled}}_{i,j}} = (\shared{x_{i, j}}, \dots,  \shared{x_{i+k, j+k}})$
    }
    \For{$i \in [0,n^2-1]$}{
        $\shared{y_i} \leftarrow \langle \shared{\vec x_i} , \mathsf{Argmax}(\shared{\vec x_i}) \rangle$\\
        where $\shared{\vec x_i} = \shared{x_{i, 0}}, \dots, \shared{x_{i, k^2}}$
    }
    \KwRet $\vec y$ reshaped as a $n \times n$ matrix.
\caption{$\mathsf{MaxPool}$ functionality using FSS. $k$ is the kernel size, the stride is fixed to 2, padding to 0 and dilation to 1.}
\label{algo:maxpool}
\end{algorithm}

\vspace*{.111cm}
\noindent
\textbf{BatchNorm}
is implemented using Newton's method as in \cite{wagh2020falcon} to implement the square inverse of the variance, as computing batch normalization exactly in a private way is very costly \cite{wagh2018securenn}. Given an input $\vec x = (x_0, \dots, x_{m-1})$ with mean $\mu$ and variance $\sigma^2$, we return $\smash{\gamma \cdot \hat \theta \cdot (\vec x - \mu) + \beta}$. Variables $\gamma$ and $\beta$ are learnable parameters and $\smash{\hat \theta}$ is the estimate inverse of $\smash{\sqrt{\sigma^2 + \epsilon}}$ with $\epsilon \ll 1$ and is computed iteratively as such: 

\[ \theta_{i+1} = \theta_i \cdot \dfrac{(C + 1) - (\sigma^2 + \epsilon) \cdot \theta_i^2}{C}  \] 

Compared to \cite{wagh2020falcon}, we do not make any costly initial approximation, therefore instead of $C = 2$ which corresponds to the classic Newton's method, we use higher values of $C$ (like $C = 6$ for the intermediate layers) which can reduce the convergence speed of the method but spares the initialisation cost.

The requirements on the approximation depend whether we are doing training or evaluation. If we are evaluating a pre-trained secret-shared neural network, having a very precise approximation is crucial, especially if the model is deep like ResNet18. Indeed, the deeper the model is, the more errors in the BatchNorm layers will propagate in the model and make it unusable. However, if the model has a running mean and variance which is the default for PyTorch, we only need to compute once the square inverse of the running variance at the beginning of the computation. 

For training however, we can use less precise approximations, since the goal of the batch normalization layer is to normalize the signal and this does not need to be done exactly as we show. We have found it very useful to reuse the result of the computation on the previous batch as an initial guess for the next batch. Moreover, we observe that for deep networks such as ResNet18, we can reduce the number of iterations of the Newton method from 4 to only 3 compared to \cite{wagh2020falcon}, except of the first batch (which does not have a proper initialisation), and for the initial and last BatchNorm layers. For those layers, which either suffer from a too high or too low variance, we increase the number of iterations. For all layers, typical relative error never exceeds $5\%$ and moderately affects learning capabilities, as our analysis on ResNet18 shows in Table \ref{table:batchnorm}. We train the model on the Hymenoptera binary classification task\footnote{https://download.pytorch.org/tutorial/hymenoptera\_data.zip} using different approximated BatchNorm layers for which we report the associated number of rounds per layer when computed in a private way. More details about our experiments on ResNet18 can be found in Appendix \ref{app:batchnorm}. \\

\begin{table} 
\caption{Accuracy of training ResNet18 on the Hymenoptera classification task with exact or approximated BatchNorm (BN)\label{table:batchnorm}}

\begin{center}
\begin{tabular}{l l c c c}
\hline
 BatchNorm$\!\!\!\!\!$ & init. with  & Newton   & Accuracy & Average comm. \\ 
      & last batch   & iterations &           & rounds per BN\\ 
\hline
 Exact & - & - &  93.59  & - \\  % 0.934641, 0.947712, 0.928105, 0.941176, 0.928105
 Approx. & True & 3 & 89.15 & 9 \\  % 0.888889, 0.921569, 0.862745, 0.895425, 0.888889
 Approx. & False & 20 & 88.24 & 60 \\  % 0.908497, 0.875817, 0.856209, 0.882353, 0.888889
 % Approximated & True & 20 & 87.71 & 60 \\ % 0.869281, 0.882353, 0.869281, 0.862745, 0.901961
 Approx. & False & 10 & 84.97 & 30 \\  % 0.856209, 0.856209, 0.830065, 0.856209, 0.849673
 Approx. & False & 3 & 60.13 & 9 \\  
 \hline
\end{tabular}
\end{center}
\end{table}

% However, it is a limitation for using pre-trained networks: we observed on AlexNet adapted to CIFAR-10 that training the model with a standard BatchNorm and evaluating it with our approximation resulted in poor results, so we had to train it using the approximated layer. Similarly, if a model was trained with an approximated BatchNorm, the same approximation should be used at evaluation.

\begin{table*} 
\caption{Theoretical online communication complexity of our protocols. Input sizes into brackets are those of the layers' parameters, where $k$ stands for the kernel size and $s$ for the stride. Communication is given in number of values transmitted, and should be multiplied by their size (typically 4 bytes). Missing entries mean that data was not available. \label{table:complexity}}

\vspace*{-.111cm}

\begin{center}
\begin{tabular}{M c c c c c c c}
\hline
 Protocol & Input size & \multicolumn{3}{c}{Rounds~~~~~~} & \multicolumn{3}{c}{~~~~~Online Communication}\\ 
 & & Ours & FALCON \cite{wagh2020falcon} & \!\!\!\!$\textrm{ABY}^3$ \cite{mohassel2018aby3}\!\! & Ours & FALCON \cite{wagh2020falcon} &  \!\!$\textrm{ABY}^3$ \cite{mohassel2018aby3}\!\! \\
\hline
 Equality & $m$ & 1 & - & 2 & $m$ & - & $\sim \lambda m$  \\  
 Comparison & $m$ & 1 & 7 & 2 & $m$ & $2m$ & $\sim \lambda m$  \\ 
 MatMul & \!\!\!$m_1 \times m_2, m_2 \times m_3$\!\!\! & 1 & 1 & 1 & $m_1 m_2 + m_2 m_3$ &  $m_1 m_3$ &  $m_1 m_3$  \\ 
 \hline
 Linear & \!\!\!\!\small{$m_1 \times m_2, \{m_2 \times m_3\}$}\!\!\!\! & 1 & 1 & - & $m_1 m_2 + m_2 m_3$ & $m_1 m_3$ & - \\ 
 Convolution & $m \times m, \{k, s\}$ & 1 & 1 & - & \!\!\!\small{$((m - k)/s+1) ^2k^2 + k^2$}\!\!\! & $\sim m^2k^2$& -\\ 
 ReLU & $m$ & 2 & 10 & - & $3m$ & $4m$ & - \\
 Argmax & $m$ & 2 & - & -& $m^2$ & -& -\\ 
 MaxPool & $m \times m , \{k, s\}$ & 3 & \small{$12(k^2-1)$} & - & \!\!\!\small{$((m - k)/s+1) ^2 (k^4  + 2)$}\!\!\! & $\sim 5m^2$ & -\\ 
 BatchNorm & $m \times m$ & 9 & 335 & - & $18m^2$ & $\sim 56m^2$ & -\\ % 1 + 1 * 0 + 2 * 3 + 2 \\
 \hline
\end{tabular}

\vspace*{-.322cm}
\end{center}
\end{table*}

Table \ref{table:complexity} summarizes the online communication cost of each ML operation presented above, and shows that basic operations such as comparison have a very efficient online communication. We also report results from \cite{wagh2020falcon} which achieve good experimental performance.

\subsection{Training Phase using Autograd}

These operations are sufficient to evaluate real world models in a fully private way. To also support private training of these models, we need to perform a private backward pass. As we overload operations such as convolutions or activation functions, we cannot use the built-in autograd functionality of PyTorch. Therefore, we have used the custom autograd functionality of the PySyft library \cite{ryffel2018generic}, where it should be specified how to compute the derivatives of the operations that we have overloaded. Backpropagation also uses the same basic blocks than those used in the forward pass, including our private comparison protocol. Therefore, the training procedure $\textsf{Train}$ described in Algorithm \ref{algo:train} closely follows the steps of plaintext training, except that the interactions between the secret shared data and model parameters use the protocols we have described in Section \ref{sec:common_ml_op}.

% Weird LateX error (but compiles)
\begin{algorithm}
    \KwIn{$\shared{x}, \shared{y_{\textsf{real}}}, \shared{\theta}$}
    \KwOut{$\shared{\hat \theta}$}
    $\textsf{opt} = \textsf{Optim}(\shared{\theta})$\\
    $\shared{y_{\textsf{pred}}} = \textsf{Forward}(\shared{\theta}, \shared{x})$\\
    $\ell = \mathcal{L}(\shared{y_{\textsf{pred}}}, \shared{y_{\textsf{real}}})$\\
    $\shared{\nabla \theta} = \textsf{Backward}(\ell, \shared{\theta})$\\
    $\shared{\theta} = \textsf{opt}(\shared{\nabla \theta} ,\shared{\theta})$\\
    \KwRet $\shared{\theta}$
\caption{$\mathsf{Train}$ procedure, that uses data $x$ to update the model $\theta$. Lines $2-5$ often run on batches extracted from $\shared{x}$ and hence are iterated until $\shared{x}$ has been completely used. $\textsf{Optim}$ refers here to the optimizer that implements (stochastic) gradient descent. $\textsf{Forward}$ and $\textsf{Backward}$ are the forward and backward passes of the model $\theta$. $\mathcal{L}$ is the loss function (mean square error or cross entropy).}
\label{algo:train}
\end{algorithm}

\vspace*{.111cm}

\iffalse % MULTI LINE COMMENT THIS IS NOT DISPLAYED
\begin{table*} 
\caption{Inference time over several popular network architectures.\label{table:inference}}

\vspace*{-.111cm}

\begin{center}
\begin{tabular}{l l c c c c}
\hline
 Model & Dataset & LAN Time (s) & $\!\!$ WAN Time (s) & $\!\!\!$ Online Comm. (MB) $\!\!\!$ & $\!\!$ Batch Size $\!\!\!\!$ \\ 
\hline
 Network-1 & MNIST      & 0.004 & 0.055 & 0.013 & 128 \\  
 Network-2 & MNIST       & 0.096 & 0.234 & 0.33 & 128 \\  
 LeNet & MNIST           & 0.129 & 0.289 & 0.46 & 128 \\  
 AlexNet & CIFAR-10      & 0.303 & 0.696 & 1.46 & 128 \\  
 AlexNet & $\!\!\!\! \small{64\!\times\!64}$ ImageNet$\!\!\!\!\!\!\!\!\!\!\!\!\!\!\!\!\!$ & 0.723 & 1.36 & 2.56 & 128 \\ 
 VGG16 & CIFAR-10        & 4.50 & 8.14 & 16.1 & 32 \\  
 %VGG16 & CIFAR-10        & 3.88 & 6.59 & 13.9 & 60 \\  
 VGG16 & $\!\!\!\! \small{64\!\times\!64}$ ImageNet$\!\!\!\!\!\!\!\!\!\!\!\!\!\!\!\!\!$   & 11.4 & 35.9 & 102 & 16 \\ 
 ResNet18 & $\!\!\!\!\!\!\small{224\!\times\!224}$ ImageNet$\!\!\!\!\!\!\!\!\!\!\!\!\!\!\!\!\!$ & 43.3 & 80.8 & 220 & 4 \\ 
 \hline
\end{tabular}

\vspace*{-.322cm}

\end{center}
\end{table*}
\fi

\section{Extension to Private Federated Learning}\label{section:n_party}
This 2-party protocol between a model owner and a data owner can be extended to an $n$-party federated learning protocol where several \emph{clients} contribute their data to a model owned by an orchestrator \emph{server}. We assume that the clients have the same set of features but have different samples in their data sets. This approach is sometimes called \textit{Horizontal Federated Learning} and is used widely, like in secure aggregation \cite{bonawitz2017practical}. 
The idea is that the server sends a version of the model to all clients, so that all clients start training the same model in parallel using their own data. With a frequency that varies depending on the settings, the server aggregates the models produced by each clients and sends back the aggregated version to be further trained by all clients. This way, clients \textit{federate} their effort to train a global model, without sharing their data. Compared to secure aggregation \cite{bonawitz2017practical}, we are less concerned with parties dropping before the end of the protocol (we consider institutions rather than phones), and we do not reveal the updated model at each aggregation or at any stage, hence providing better privacy.

Algorithm \ref{algo:federatedFSS} shows one possible implementation of fully private federated learning using 2-party function secret sharing. It prevents collusion between at most $k$ out of $n$ clients, the threat being that a client receiving the share of another client during aggregation phase could collude with the server to help reconstructing the model contributed by this client, and infer information about its private data. This aggregation requires extra communication rounds but this is in practice negligible compared to the training procedure $\textsf{Train}$ initiated between a server and a client. Note that other aggregation mechanisms could be used, including using n-party MPC protocols or homomorphic encryption, but we proposed masking as this is quite in line with the concept of FSS where we mask the private input with $\alpha$.

% At the beginning of the interaction, the server and model owner initializes its model and builds $n$ pairs of additive shares of the model parameters. For each pair $i$, it keeps one of the shares and sends the other one to the corresponding client $i$. Then, the server runs in parallel the training procedure with all the clients until the aggregation phase starts. Aggregation for the server shares is straightforward, as the $n$ shares it holds can be simply locally averaged. But the clients have to average their shares together to get a client share of the aggregated model. One possibility is that clients broadcast their shares and compute the average locally. However, to prevent a client colluding with the server from reconstructing the model contributed by a given client, they hide their shares using masking. 
% This can be done using correlated random masks: client $i$ generates $k$ seed and sends them to the clients $i+1$ to $i+k$, one to each client, while receiving $k$ seeds from client $i-k$ to $i-1$. Client $i$ then generates $k$ random masks $(M_{j})_{j=1..k}$ using its seeds and $k$ other masks $(\hat M_j)_{j=1..k}$ using the ones received. It then publishes its share, masked with $\sum_{j=1..k} M_j - \hat M_j$. As the masks cancel each other out, the computation will be correct. In addition, this scheme is secure against a collusion between the server and $k$ clients (with $k < n$), since no other client other than $i$ knows more than one mask $M_j$ and one mask $\hat M_j$.

% \begin{algorithm*}[!htb]
% Weird LateX error (but compiles)
\begin{algorithm*}[!htb]

    \KwIn{Model on the server $S$, initialized with parameters $\theta$}
    \KwOut{Model trained using the data from the clients $(C_i)_{i=1..n}$}
    \textbf{Initialisation} $S$ secret shares $\theta$ with the clients. \\
    $\shared{\theta} \leftarrow_{\textsf{share}} \theta$ \\ 
    \For{$i \in [1,n]$}{
        $\shared{\theta_i} \leftarrow_{\textsf{copy}} \shared{\theta}$ \\
        S stores $\shared{\theta_i}_0$ and sends $\shared{\theta_i}_1$ to $C_i$ \\
    }
    \textbf{Training} $S$ runs in parallel $n$ training procedures. \\
    \For{$i \in [1,n]$}{
        $\shared{\hat \theta_i} \leftarrow \textsf{Train}(\shared{x_i}, \shared{y_i}, \shared{\theta_i})$, where $\shared{x_i}$ and $\shared{y_i}$ are the data and corresponding labels from $C_i$
    }
    \textbf{Aggregation} All updated models are aggregated with a scheme secure against collusion between the server and $k$ clients. \\
    $S$ computes $\shared{\hat{\theta}}_0 := \sum_{i=1..n} \shared{\hat \theta_i}_0$\\
    % $C^*$ is elected among $(C_i)_{i=1..n}$, at random or on a rotating basis. \\
    $C^* \overset{{\scriptscriptstyle\$}}{\leftarrow} (C_i)_{i=1..n}$ \\
    \For{$i \in [1,n]$}{
        $C_i$ generates $k$ seeds and sends them to $C_{i+1 \mod n}$, $\dots$, $C_{i+k \mod n}$ \\
        $C_i$ receives $k$ seeds from $C_{i-k \mod n}$, $\dots$, $C_{i-1 \mod n}$ \\
        $C_i$ derives $k$ random masks $(m_j)_{j=1..k}$ from its own seeds \\
        $C_i$ derives $k$ random masks $(\hat{m}_j)_{j=1..k}$ from the seeds received \\
        $C_i$ builds a global mask $\mu_i = \sum_{j=1..k} m_j - \hat{m}_j$ \\
        $C_i$ sends $\shared{\theta_i}_1 + \mu_i$ to $C^*$ 
    }
    $C^*$ receives $\shared{\hat{\theta}}_1 := \sum_{i=1..n} \shared{\theta_i}_1 + \sum_{i=1..n} \mu_i = \sum_{i=1..n} \shared{\theta_i}_1$. \\
    $C^*$ broadcasts $\shared{\hat{\theta}}_1$ to all clients. \\
    Iterate \textbf{Training} and \textbf{Aggregation} using $\shared{\hat{\theta}}$ until the training is complete. \\
    \KwRet $\shared{\hat{\theta}}$
\caption{Federated Learning algorithm using 2-party Function Secret Sharing}
\label{algo:federatedFSS}
\end{algorithm*}

\vspace*{.111cm}

%This can be done using correlated random masks: client $i$ generates a seed, sends it to the clients $i+1$ to $i+k$ while receiving one from client $i-1$. Client $i$ then generates a random mask $M_i$ using its seed and another $M_{i-1}$ using the one of client $i-1$ and publishes its share masked with $M_i - M_{i-1}$. As the masks cancel each other out, the computation will be correct.

\section{Experiments} \label{section:experiments}
In order to simplify comparison with existing work, we follow a setup very close to the work of  \cite{wagh2020falcon}. The reason why we compare our work to \cite{wagh2020falcon} is that it provides the most extensive experiments of private training and evaluation we are aware of. We are aware that \cite{wagh2020falcon} also provides honest-majority malicious security, but we only report their results in the honest but curious setting (where they obtain the best runtimes). We assess private inference of several networks on the datasets MNIST \cite{lecun2010mnist}, CIFAR-10 \cite{krizhevsky2014cifar}, 64$\times$64 Tiny Imagenet \cite{wu2017tiny, imagenet_cvpr09} and 224$\times$224 Hymenoptera which is a subset of Imagenet, and we also benchmark private training on MNIST. More details about the datasets used can be found in Appendix \ref{app:datasets}. More precisely, we assess 6 networks: a 3 layers fully-connected network (Network-1), a small convolutional network with maxpool (Network-2), LeNet \cite{lecun1998gradient}, AlexNet \cite{krizhevsky2012alexnet}, VGG16  \cite{simonyan2014vgg} and ResNet18 \cite{he2016resnet} which to the best of our knowledge has never been studied before in private deep learning. The description of these networks is available in Appendix \ref{appendix:NNarchi}.

Our implementation provides a Python interface and is tightly coupled with PyTorch to provide both the ease of use and the expressiveness of this library. To use our protocols that only work in finite groups like  $\Z_{2^{32}}$, we convert our input values and model parameters to fixed precision. To do so, we rely on the PySyft library \cite{ryffel2018generic} which extends common deep learning frameworks including PyTorch with a communication layer for federated learning and supports fixed precision. The experiments are run on Amazon EC2 using m5d.4xlarge machines for CPU benchmarks and g4dn.4xlarge for GPU, both with 16 cores and 64GB of CPU RAM, and we report our results both in the LAN and in the WAN setting. Latency is of 70ms for the WAN setting and is considered negligible in the LAN setting. Last, all values are encoded on 32 bits.

\begin{table*} 
\caption{Comparison of the inference time between secure frameworks on several popular neural network architectures. FALCON, SecureNN, and $\smash{\textrm{ABY}^3}$ are 3-party protocols. XONN and Gazelle are 2-party protocols. All protocols are evaluated in the honest-but-curious setting. 
For each network we report in order 
the runtime for computing the preprocessing (if any) using CPUs, 
the runtime for the online phase in LAN using CPUs, 
the runtime for the online phase in LAN using GPUs,
the runtime for the online phase in WAN using CPUs, 
and the communication needed during the online phase. Runtime is given in seconds and communication in MB.  Missing entries mean that data was not available.
\label{table:inference}}
\vspace*{-.111cm}
\begin{center}
\begin{tabular}{ l r | c c c c c | c c c c c | c c c c c }
\hline
  & & \multicolumn{5}{c|}{Network-1} & \multicolumn{5}{c|}{Network-2} & \multicolumn{5}{c}{LeNet}\\ 
  & & \!\!\!LAN\!\!\! & \!\!\!LAN\!\!\! & \!\!\!LAN\!\!\! & \!\!\!WAN\!\!\! 
  & & \!\!\!LAN\!\!\! & \!\!\!LAN\!\!\! & \!\!\!LAN\!\!\! & \!\!\!WAN\!\!\! 
  & & \!\!\!LAN\!\!\! & \!\!\!LAN\!\!\! & \!\!\!LAN\!\!\! & \!\!\!WAN\!\!\! & \\
Framework & \!\!\!\!\!\!Dataset\!\!
    & \!\!\!Prep.\!\!\! & \!\!\!CPU\!\!\! & \!\!\!GPU\!\!\! & \!\!\!CPU\!\!\! &  \!\!\!\!\!Comm.\!\!\!\!\!
    & \!\!\!Prep.\!\!\! & \!\!\!CPU\!\!\! & \!\!\!GPU\!\!\! & \!\!\!CPU\!\!\! &  \!\!\!\!\!Comm.\!\!\!\!\!
    & \!\!\!Prep.\!\!\! & \!\!\!CPU\!\!\! & \!\!\!GPU\!\!\! & \!\!\!CPU\!\!\! &  \!\!\!\!\!Comm. \\
\hline
AriaNN    & \!\!\!\!\!\!\!\!MNIST\!\! 
                  & \!\!0.002\!\! & \!\!0.004\!\! & \!\!0.002\!\! & \!\!0.043\!\! & \!\!0.022\!\!  
                  & \!\!0.028\!\! & \!\!0.041\!\! & \!\!0.024\!\! & \!\!0.133\!\! & \!\!0.28\!\! 
                  & \!\!0.041\!\! & \!\!0.055\!\! & \!\!0.035\!\! & \!\!0.143\!\! & \!\!0.43\!\! \\
FALCON    & \!\!\!\!\!\!\!\!MNIST\!\! 
                  & -     & \!\!\!0.011\!\!\! & -     & \!\!\!0.990\!\!\! & \!\!\!0.012\!\!\!
                  & -     & \!\!\!0.009\!\!\! & -     & \!\!\!0.760\!\!\! & \!\!\!0.049\!\!\!
                  & -     & \!\!\!0.047\!\!\! & -     & \!\!\!3.06\!\!\!  & \!\!\!0.74\!\!\! \\
SecureNN  & \!\!\!\!\!\!\!\!MNIST\!\! 
                  & -     & \!\!\!0.043\!\!\! & -     & \!\!\!2.43\!\!\!  & \!\!\!2.1\!\!\!
                  & -     & \!\!\!0.130\!\!\! & -     & \!\!\!3.93\!\!\!  & \!\!\!8.86\!\!\!
                  & -     & -     & -     & -     & -    \\
XONN      & \!\!\!\!\!\!\!\!MNIST\!\!
                  & -     & \!\!\!0.130\!\!\! & -     & -     & \!\!\!4.29\!\!\!
                  & -     & \!\!\!0.150\!\!\! & -     & -     & \!\!\!32.1\!\!\!
                  & -     & -     & -     & -     & -    \\
Gazelle   & \!\!\!\!\!\!\!\!MNIST\!\! 
                  & \!\!\!0\!\!\!  & \!\!\!0.030\!\!\! & -     & -     & \!\!\!0.5\!\!\! 
                  & \!\!\!0.481\!\!\! & \!\!\!0.330\!\!\! & -     & -     & \!\!\!22.5\!\!\!
                  & -     & -     & -     & -     & -    \\
$\smash{\textrm{ABY}^3}$& \!\!\!\!\!\!\!\!MNIST\!\! 
                  & \!\!\!0.005\!\!\! & \!\!\!0.003\!\!\! & -     & -     & \!\!\!0.5 \!\!\! 
                  & -     & -     & -     & -     & -   
                  & -     & -     & -     & -     & -    \\
CrypTFlow  & \!\!\!\!\!\!\!\!MNIST\!\! 
                  & -     & \!\!\!0.008\!\!\! & -     & -     & -    
                  & -     & \!\!\!0.034\!\!\! & -     & -     & -    
                  & -     & \!\!\!0.058\!\!\! & -     & -     & -    \\
\hline
\end{tabular}

\vspace*{0.1cm}

\begin{tabular}{ l r | c c c c c | c c c c c | c c c c c }
\hline
  & & \multicolumn{5}{c|}{AlexNet} & \multicolumn{5}{c|}{VGG16} & \multicolumn{5}{c}{ResNet18}\\ 
  & & \!\!\!LAN\!\!\! & \!\!\!LAN\!\!\! & \!\!\!LAN\!\!\! & \!\!\!WAN\!\!\!
  & & \!\!\!LAN\!\!\! & \!\!\!LAN\!\!\! & \!\!\!LAN\!\!\! & \!\!\!WAN\!\!\!
  & & \!\!\!LAN\!\!\! & \!\!\!LAN\!\!\! & \!\!\!LAN\!\!\! & \!\!\!WAN\!\!\! & \\
Framework & Dataset  
    & \!\!\!Prep.\!\!\! & \!\!\!CPU\!\!\! & \!\!\!GPU\!\!\! & \!\!\!CPU\!\!\! &  \!\!\!\!\!Comm.\!\!\!\!\!
    & \!\!\!Prep.\!\!\! & \!\!\!CPU\!\!\! & \!\!\!GPU\!\!\! & \!\!\!CPU\!\!\! &  \!\!\!\!\!Comm.\!\!\!\!\!
    & \!\!\!Prep.\!\!\! & \!\!\!CPU\!\!\! & \!\!\!GPU\!\!\! & \!\!\!CPU\!\!\! &  \!\!\!\!\!Comm.  \\
\hline
AriaNN & CIFAR-10\!\!
                  & \!\!\!0.091\!\!\! & \!\!\!0.15\!\!\!  & \!\!\!0.078\!\!\! & \!\!\!0.34\!\!\!  & \!\!\!0.95\!\!\!
                  & \!\!\!0.94\!\!\!  & \!\!\!1.75\!\!\!  & \!\!\!1.55\!\!\!  & \!\!\!1.99\!\!\!  & \!\!\!12.59\!\!\!
                  & -     & -     & -     & -     & -    \\
FALCON & CIFAR-10\!\! 
                  & -     & \!\!\!0.043\!\!\! & -     & \!\!\!0.13\!\!\!  & \!\!\!1.35\!\!\!
                  & -     & \!\!\!0.79\!\!\!  & -     & \!\!\!1.27\!\!\!  & \!\!\!13.51\!\!\!
                  & -     & -     & -     & -     & -    \\
AriaNN & $\!\!\!\!\!\!\!\!\small{64\!\times\!64}$ ImageNet\!\!
                  & \!\!\!0.27\!\!\!  & \!\!\!0.33\!\!\!  & \!\!\!0.20\!\!\!  & \!\!\!0.48\!\!\!  & \!\!\!1.75\!\!\!
                  & \!\!\!3.42\!\!\!  & \!\!\!7.51\!\!\!  & \!\!\!6.83\!\!\!  & \!\!\!8.00\!\!\!  & \!\!\!53.11 \!\!\!
                  & -     & -     & -     & -     & -    \\
FALCON & $\!\!\!\!\!\!\!\!\small{64\!\times\!64}$ ImageNet\!\!
                  & -     & \!\!\!1.81\!\!\!  & -     & \!\!\!2.43\!\!\!  & \!\!\!19.21\!\!\!
                  & -     & \!\!\!3.15\!\!\!  & -     & \!\!\!4.67\!\!\!  & \!\!\!52.56\!\!\!
                  & -     & -     & -     & -     & -    \\
AriaNN & $\!\!\!\!\!\!\!\!\!\!\!\!\!\!\!\small{224\!\times\!224}$ Hymenoptera\!\!
                  & -     & -     & -     & -     & -    
                  & -     & -     & -     & -     & -     
                  & \!\!\!10.02\!\!\! & \!\!\!19.88\!\!\! & \!\!\!13.90\!\!\! & \!\!\!24.07\!\!\! & \!\!\!148\!\!\! \\
                  
\hline
\end{tabular}

%% CPU
% Resnet18: BS = 8
% VGG16 CIFAR10: 64
% VGG16 Tiny Imagenet: 16 (all 32 -> 7.1s ; all 64 -> 6.48)

%% GPU
% Alexnet Tiny Imagenet BS = 64
% VGG16 CIFAR10: batch_size = 14
% VGG16 Tiny Imagenet: BS = 3 
% ResNet18 GPU BS = 1

% GPU Alex TINYI BS=32 with prep.
%  LAN ->      
%  WAN Oms ->  0.2312
%  WAN 70ms -> 1.2528

% GPU Network2 MNIST BS=128 with prep.
% LAN ->      0.024
% WAN Oms  -> 0.029
% WAN 70ms -> 0.140

\vspace*{-.322cm}
\end{center}
\end{table*}

\subsection{Inference Time and Communication} 

Comparison of experimental runtimes should be taken with caution, as different implementations and hardware may result in significant differences even for the same protocol. 
We report our online inference runtimes in Table \ref{table:inference} and show that they compare favourably with existing work including \cite{liu2017oblivious, mohassel2018aby3, mohassel2017secureml, wagh2018securenn, wagh2020falcon}. For example, 
our CPU implementation of Network-1 outperforms all other studied frameworks by at least a factor $2\times$ in the LAN setting and even more in the WAN setting. 
For larger networks such as AlexNet and VGG16, we have an execution time which is slightly higher than \cite{wagh2020falcon}. One reason for this can be that we use a Python interface to serialize messages and communicate between parties, while \cite{wagh2020falcon} uses exclusively C code.
However, we are more communication-efficient than \cite{wagh2020falcon} for models starting from LeNet, with a typical gain of 7\% to 30\% on CIFAR-10. 
Regarding the high advantage we have on AlexNet and 64$\times$64 Tiny Imagenet, this is explained by the fact that \cite{wagh2020falcon} uses a modified and more complex AlexNet while we use the one from PyTorch. Details about our networks architecture is given in Appendix \ref{appendix:NNarchi_detailed}.

Results are given for a batched evaluation with a default batch size of 128 to amortize the communication cost, as in other works compared here. For larger networks, we reduce the batch size to have the preprocessing material (including the function keys) fitting into RAM, which reduces the benefit of amortization. The exact values chosen are available in Appendix \ref{appendix:extended_inference}.

We have also added the results of our GPU implementation, which offers a clear speed-up compared to CPU with an execution which is between 10\% and 100\% faster. While this already shows the usefulness of using GPUs, one could expect a greater speed-up. One reason is that classic GPUs currently offer 16GB of RAM, which is a clear limitation for our work where we store keys in RAM. Storing the keys on the CPU would come at a marginal cost of importing them on the GPU during the online phase but would allow to use bigger batches and hence better amortize the computation.

\begin{table} 
\caption{Accuracy of pre-trained neural network architectures, evaluated over several datasets in plaintext, fixed precision and privately using FSS. Time for private evaluation in the LAN setting is also reported. \label{table:private_accuracy}
}

\vspace*{-.111cm}

\begin{center}
\begin{tabular}{l l c c c c}
\hline
  &  & LAN & \multicolumn{3}{c}{Accuracy}  \\
 Model & Dataset & $\!\!\!\!\!$ time (h)$\!\!\!\!\!$ & Private  & $\!\!\!\!\!\!$Fix prec.$\!\!\!\!\!\!$  & Public  \\ 
\hline
 %               Time & Priv & FixP & Pub
 Network-1 & MNIST  & 0.01 & 98.2 & 98.2 & 98.2 \\  
 Network-2 & MNIST  & 0.18 & 99.0 & 99.0 & 99.0 \\  
 LeNet & MNIST      & 0.24 & 99.2 & 99.3 & 99.3 \\  
 AlexNet & CIFAR-10 & 0.60 & 70.3 & 70.3 & 70.3 \\      
 AlexNet & $\!\!\!\!\!\!\!\!64\!\times\!64$ ImageNet\!\!\!\!\!\! & 0.48 & 38.3 & 38.6  & 38.6 \\  
 VGG16 & CIFAR-10 & 5.19 & 87.4 & 87.4 & 87.4\\  
 VGG16 & $\!\!\!\!\!\!\!\!64\!\times\!64$ ImageNet\!\!\!\!\!\! & 9.97 & 55.2 & 56.0 & 55.9 \\
 ResNet18 & Hymenoptera & 0.95 & 94.7 & 94.7 & 95.3 \\
 \hline
\end{tabular}

\vspace*{-.322cm}

\end{center}
\end{table}

\begin{table} 
\caption{Accuracy of neural network architectures trained over several datasets in plaintext, fixed precision and privately using FSS. Time for private training in the LAN setting is given in hours per epoch.\label{table:training}}

\vspace*{-.111cm}

%\begin{left}
\begin{tabular}{l l c c c c c}
\hline
 & & $\!\!\!\!\!$LAN time$\!\!\!\!\!$ & \multicolumn{3}{c}{Accuracy} &  \\
 Model & Dataset\!\!\!\!\! & $\!\!\!\!\!$per epoch (h)$\!\!\!\!\!$ & $\!\!$Private  & $\!\!\!\!\!\!\!\!$Fix prec.$\!\!\!\!\!\!$  & $\!\!$Public  & $\!\!\!\!\!\!$Epochs\\ 
\hline
 Network-1\!\!\!\!\! & MNIST & 0.78 & 98.0 & 98.0 & 98.2 & 15 \\  % --lr 0.01
 Network-2\!\!\!\!\! & MNIST & 2.8 & 98.3 & 99.0 & 99.0 & 10 \\ % --lr 0.02 
 LeNet\!\!\!\!\!     & MNIST & 4.2 & 99.2 & 99.2 & 99.3 & 10 \\  % --lr 0.01
 % AlexNet & CIFAR-10$\!\!\!\!\!\!\!\!$ & x & x & x & x & 15  \\ % 
 \hline
\end{tabular}

\vspace*{-.322cm}

%\end{left}
\end{table}

\subsection{Test Accuracy} 

Thanks to the flexibility of our framework, we can train each of these networks in plaintext and need only one line of code to turn them into private networks where all parameters are secret shared, or to fixed precision networks where all parameters are converted to fixed precision but computation is still in plaintext. Comparing the performance of private models with their fixed precision version helps us to understand if fixed precision by itself reduces the accuracy of the model, and gives an estimate of the loss that is related to using secret shared computation.

We compare the accuracy of several pre-trained networks in these 3 modes in Table \ref{table:private_accuracy} by running a private evaluation with FSS, a fixed precision using only PySyft and a public evaluation where the model is not modified. We observe that accuracy is well preserved overall and that converting to fixed precision has no impact on the accuracy of the model. We have a small reduction in accuracy for the two private models evaluated on $64\!\times\!64$ ImageNet but it remains close to the plaintext baseline. This gap can be explained by the fact that PySyft uses a basic and approximate private truncation after multiplication where truncation is directly applied on the shares, and by the error rate of our FSS comparison protocol. The drop in accuracy on ResNet18 is also minor and corresponds to a single mislabeled item.

If we degrade the encoding precision which by default considers values in $\Z_{2^{32}}$, or the fixed precision which is by default of 4 decimals, performance degrades as shown in Appendix \ref{appendix:precision}. 

% Note that the plaintext training is done with the same optimizer and loss and the private training, which explains why accuracies are slightly under what one would expect with those models. 

\subsection{Training Accuracy}

We have also assessed the ability of training neural networks from scratch in a private way using AriaNN. Private training is an end-to-end private procedure, which means the model or the gradients are never accessible in plaintext. We use stochastic gradient descent (SGD) with momentum, a simple but popular optimizer, and support several losses such as mean square error (used for Network-1) and cross entropy (used for Network-2 and LeNet). We report the runtime and accuracy obtained by training from scratch and evaluating several networks in Table \ref{table:training}, in plaintext, in fixed precision and in a fully private way, just as we did for inference. Note that because of the training setting, accuracy might not match best known results, but the training procedure is the same for all training modes which allows for fair comparison.

We observe that the training is done almost perfectly both in fixed precision and private mode compared to the plaintext counterpart. The only noticeable difference we observe is for Network-2, where the privately-trained model achieves 98.3\% while 99.0\% is expected. The fixed-precision accuracy which is 99.0\% suggests that our autograd functionality is working properly, so the difference must be explained by the small failure rate of FSS. Training profiles show that the accuracy starts decreasing roughly after 3 epochs, while it is supposed to keep increasing smoothly up to the 10\textsuperscript{th} epoch. Instability caused by some FSS failures could account for this behaviour. However, training on LeNet did not suffer from the same phenomenon.

Recently, \cite{fantasticfour} also reports accuracy results when training securely Network-1 on MNIST, using a 3-party semi-honest protocol that mixes \cite{mohassel2018aby3} and \cite{araki2016high}. They achieve 97.8 \% of accuracy in 15 epochs with a runtime of only 33.8s per epoch in the LAN setting. However, they do not provide a detailed comparison between the accuracy achieved with private training, and cleartext training. One major difference with our work is that we are more communication efficient. We only require 10.3MB of communication during the online phase while they use 33.8MB per epoch. In addition, they rely on $\textrm{ABY}^3$, which mean they use much more interaction rounds, which could be costly in the WAN setting although this is not monitored by this work.

%However, for models which involve convolution layers, even smaller ones such as Network-2, performances of private training are somewhat behind the plaintext baseline. In particular, the drop of performance is mainly between the plaintext model and the fixed precision one. We suspect that the custom autograd module from PySyft that we use for the backpropagation step doesn't handle convolutions correctly. To confirm this, we have opened the convolution kernel during the fixed precision training and have indeed observed that it doesn't change over time. We have reported this issue to the PySyft community. Next, to verify that this bug was the reason for the drop in accuracy, we have replicated the training of the public model but we have freezed the convolution parameters. Such a model achieves 97.2\% of accuracy instead of 99.0\%, which suggests that fixing the autograd for convolution would provide an important increase of performance, while maybe not reaching completely the plaintext accuracy. The remaining gap might be associated with the fixed precision rounding error or with additional bugs in the autograd functionality.

Training cannot complete in reasonable time for larger networks such as VGG16, which in practice might be fine-tuned rather than trained from scratch. Note that training time includes the time spent building the preprocessing material, as it is too large to be fully processed and stored in RAM in advance.

\subsection{Computation and Communication Analysis}

We have provided in Table \ref{table:compute} a small analysis of how the compute time can be decomposed. We use AlexNet on the Tiny Imagenet dataset as it is the biggest network on which we could use a batch size higher than 64 both on CPU and GPU and hence amortize the serialization and communication cost.

\begin{table}[htb]
\caption{Distribution of the compute time during inference of AlexNet on the Tiny Imagenet dataset using either CPUs or GPUs.\label{table:compute}}

\vspace*{-.111cm}

\begin{center}
\begin{tabular}{l c c c c}
\hline
  &  & MatMul and & Serialization & \\
 Processor & FSS & Convolution & and Deser. & Other \\
\hline
 CPU    &  16\%  &  72\% & 4\% & 8\%\\
  & 53ms & 238ms & 13ms & 26ms \\
 GPU & 51\% & 39\% & 8\%   & 2\% \\
    & 102ms & 78ms & 16ms & 4ms \\
 \hline
\end{tabular}

\vspace*{-.322cm}

\end{center}
\end{table}

Thanks to the efficiency of our Rust implementation, function secret sharing only accounts for 16\% for the online runtime when we use CPUs, and most of the time is spent doing matrix multiplications and convolutions. This last part uses the underlying PyTorch functions on integers which are significantly slower than when they run on floats. This motivates us to use GPUs for which such operations are far more efficient. In the GPU setting, the distribution of time is indeed much more balanced, and having function secret sharing directly running on GPUs avoids going back and forth between the CPU and the GPU.

% GPU Alexnet Tiny Imagenet BS=64 with prep.
% Total: 12.86
% FSS EVAL: 6.61
% serialize + deserialize: 1.04 = 0.66 + 0.38
% spdz_compute : 5.12 (and 5. of conv)

% CPU  Alexnet Tiny Imagenet BS = 128
% Total 43.43
% FSS EVAL: 7
% serialize + deserialize: 1.93 = 1.20 + 0.73
% spdz_compute: 31.09

Regarding the trade-off between computation and communication time, we show in Table \ref{table:comp_vs_comm} that in the WAN setting and using CPUs, computation appears to be the main bottleneck especially for bigger models. This also encourages us to further improve the GPU implementation, as any optimization of the computation efficiency will have an important impact on the overall runtime.
% We have omitted the communication time as it is highly dependent on the latency and the bandwidth of the network.

\begin{table}[htb]
\caption{Proportion of the overall runtime spent on computation versus communication, in the WAN setting using CPUs\label{table:comp_vs_comm}}

\vspace*{-.111cm}

\begin{center}
\begin{tabular}{l c c c}
\hline
 Model & Dataset & Computation (\%) & Comm. (\%) \\
\hline
 Network-1 & MNIST  & 9  & 91 \\  
 Network-2 & MNIST  & 31 & 69 \\  
 LeNet & MNIST      & 38 & 62 \\  
 AlexNet & CIFAR-10 & 44 & 56 \\      
 AlexNet & $\!\!\!\!\!\!\!\!64\!\times\!64$ ImageNet\!\!\!\!\!\! & 69 & 31 \\  
 VGG16 & CIFAR-10 & 88 & 12 \\  
 VGG16 & $\!\!\!\!\!\!\!\!64\!\times\!64$ ImageNet\!\!\!\!\!\! & 93 & 7 \\
 ResNet18 & Hymenoptera & 83 & 17 \\
 \hline
\end{tabular}

\vspace*{-.322cm}

\end{center}
\end{table}

\subsection{Discussion}

Regarding experiments on larger networks, we could not use batches of size 128. This is mainly due to the size of the comparison function keys, which is currently proportional to the size of the input tensor, with a multiplicative factor of $n\lambda$ where $n=32$ and $\lambda=128$. Optimizing the function secret sharing protocol to reduce the size of those keys would allow to better amortize batched computations and would also reduce the runtime as we would manipulate smaller arrays during the private comparison. An interesting other improvement would be to run experiments on $n=16$ bits instead of $32$. Classic ML frameworks like PyTorch or TensorFlow now support 16 bits encoding both on CPU and GPU.

We have proposed a first implementation of FSS on GPU, which can still be improved to reduce the memory footprint of they keys. Further efforts could be made to decrease it roughly by 50\% to match the theoretical key size. In addition, and as the small difference between the LAN and the WAN runtime shows, especially for bigger networks, most of the time is now spent on computation. Therefore, optimizing computation on GPUs will have a direct impact on the overall efficiency of the inference or the training.

We have shown the relevance of using FSS for private training and evaluation of models in machine learning. Compared to concurrent works like \cite{boylefunction2020}, we have shown that we have very competitive protocols, and that the failure rate of the comparison protocol has no impact for machine learning applications. Our protocol has been used in one recent work \cite{kaissis2021end} where it was applied to the field of medical imaging on chest X-rays.

%Last, training accuracies presented in Table \ref{table:training} do not match state-of-the-art performance for the models and datasets considered. While some improvements clearly need to be made on the autograd system that we use, there will still be a gap for bigger models as we use a simplified training procedure. Supporting losses such as cross entropy, more complex optimizers like SGD with momentum or Adam, and dropout layers would be an interesting follow-up to improve our training performance.

\section{Conclusion}
In this work, we improve over the best known protocols for private comparison using function secret sharing by reducing the keys size by almost a factor $\times4$. We show how this new algorithm helps us implement efficient machine learning components and we provide constructions for ReLU and MaxPool with only 2 and 3 rounds of communication. Additionally, we show that AriaNN can implement a large diversity of neural networks, from convolutional networks to ResNet18, which are very competitive in terms of runtime and communication compared to existing work.
%which outperform the ones of efficient private ML frameworks like \cite{wagh2020falcon}. Additionally, we propose an approximate and inexpensive construction for BatchNorm that can be used to accelerate training without damaging significantly performance. 
Last, we provide an implementation of AriaNN which can run both on CPU and GPU, providing promising runtime improvements for the next generation of hardware accelerated privacy-preserving machine learning models.

\section*{Acknowledgments}

We would like to thank Geoffroy Couteau, Chloé Hébant and Loïc Estève for helpful discussions throughout this project. We are also grateful for the long-standing support of the OpenMined community and in particular its dedicated cryptography team, including George Muraru, Rasswanth S, Hrishikesh Kamath, Arturo Marquez, Yugandhar Tripathi, S P Sharan, Muhammed Abogazia, Alan Aboudib, Ayoub Benaissa, Sukhad Joshi and many others.

This work was supported in part by the French project FUI ANBLIC. The computing power was graciously provided by the French company ARKHN.

\bibliographystyle{plain}
\bibliography{ref}

\newpage

\appendix
% Put back in the core paper
\section{FSS Comparison Protocol - Security Proof}\label{appendix:security}
For this proof, we follow the same process than \cite{boyle2016function}.

\noindent We prove that each party's key $\mathsf{k}_j$ is pseudorandom. This is done via a sequence of hybrid distributions, where in each step we replace two correction words $\CWi$ and $\CWi_{leaf}$ within the key from being honestly generated to being random. In the initial game, all the correction words are as in the real distribution, and in the last game, they are all random. As every gaps are indistinguishable for any polynomially-bounded adversary, the real distribution is indistinguishable from random: this proves the pseudo-randomness of the keys.

The high-level argument for security will go as follows. Each party $j \in \zerone$ begins with a share $\shared{\alpha}_j$ and a random seed $s^{(1)}_j$ that are completely unknown to the other party. In each level of key generation (for $i = 1$ to $n$), the parties apply a PRG to their seed $s^{(i)}_j$ to generate 8 items: namely, 2 seeds $s_j^\mathsf{L} , s_j^\mathsf{R} $, 2 bits $t_j^\mathsf{L} , t_j^\mathsf{R} $, 2 $n$-bits values $\sigma_j^\mathsf{L}, \sigma_j^\mathsf{R}$ and 2 other bits $\tau_j^\mathsf{L}, \tau_j^\mathsf{R}$. This process is always done on a seed which appears completely random given the view of the other party. Hence, the security of the PRG guarantees that the 8 resulting values appear similarly random given the view of the other party. The $i$th level correction word $\CWi$ will “use up” the secret randomness of 3 of the 4 first pieces: the two bits $t_j^\mathsf{L} , t_j^\mathsf{R}$, and the seed $s_j^\mathsf{Lose}$ corresponding to the direction exiting the special path i.e. $\mathsf{Lose} = \mathsf{L}$ if $\alpha[i] = 1$ and $\mathsf{Lose} = \mathsf{R}$ if $\alpha[i] = 0$. However, given this $\CWi$, the remaining seed $s_j^\mathsf{Keep}$ for $\mathsf{Keep} \neq \mathsf{Lose}$ is still unpredictable to the other party, as it is kept hidden.
Similarly, the $i$th level correction word $\CWi_{leaf}$ uses up the secret randomness of the 4 last pieces, $\sigma_j^\mathsf{L}, \sigma_j^\mathsf{R}$ and $\tau_j^\mathsf{L}, \tau_j^\mathsf{R}$, and appears random given the view of the other party. 
The argument is then continued in similar fashion to the next level, which uses $s_j^\mathsf{Keep}$ as an input to the PRG. 

For each $i \in \{0,1, \dots ,n+1\}$, we will consider a hybrid distribution $\mathsf{Hyb}_i$ defined roughly as follows, for $j \in \zerone$:
\begin{enumerate}
 \item $s_j^{(1)} \sampled \zerone^\lambda$ chosen at random (honestly), and $t_j^{(1)} = j$.
 \item $CW^{(1)}, \dots, CW^{(i)} \leftarrow \zerone^{2(\lambda + n + 2)}$  and $CW_{leaf}^{(1)}, \dots, CW_{leaf}^{(i)} \leftarrow \zerone^{n}$ chosen at random.
 \item For $k < i$, $s_j^{(k+1)}||t_j^{(k+1)}, \sigma_j^{(k+1)}||\tau_j^{(k+1)}$ computed honestly, as a function of $s_j^{(0)} \sep t_j^{(0)}$ and $CW^{(1)}, \dots, CW^{(k)}$.
 \item For $i$, the other party’s seed $s_{1-j}^{(i)} \leftarrow \zerone^\lambda$ is chosen at random, $t_{1-j}^{(i)} = 1 - t_{j}^{(i)}$, $\sigma_{1-j}^{(i)} = \sigma_{j}^{(i)}$, and $\tau_{1-j}^{(i)} = \tau_{j}^{(i)}$.
 \item For $k \ge i$: the remaining values $s_j^{(k+1)} \sep t_j^{(k+1)}$ , $s_{1-j}^{(k+1)}  \sep t_{1-j}^{(k+1)}$ , $CW^{(k)}$, $\sigma_j^{(k+1)} \sep \tau_j^{(k+1)}$ , $\sigma_{1-j}^{(k+1)}  \sep \tau_{1-j}^{(k+1)}$ , $CW_{leaf}^{(k)}$ all computed honestly, as a function of the previously chosen values.
 \item The output of the experiment is $\key_j := \shared{\alpha}_j \sep  s_j^{(1)} \sep (CW^{(i)})_{i=1..n} \sep (CW_{leaf}^{(i)})_{i=1..n+1}$.
\end{enumerate}

$\mathsf{Hyb}_i$ is formally described in Algorithm \ref{algo:hyb}. When $i = 0$, the algorithm corresponds to the honest key generation, while when $i = n+1$, it generates a completely random key. We only need to prove that for any $i \in \{1, \dots n+1\}$, $\mathsf{Hyb}_{i-1}$ and $\mathsf{Hyb}_i$ are indistinguishable based on the security of our PRG.

More precisely, let us first consider $i \le n$.

\begin{claim}\label{claim:hyb_i}
    There exists a polynomial $p'$ such that for any $(T,\epsilon_{\textsf{PRG}})$-secure pseudorandom generator $G$, then for every  $i \le n$, $j \in \zerone$, and every non-uniform adversary $\mathcal{A}$ running in time $T' \le T - p' (\lambda)$, it holds that
\[
         \big|  \textrm{Pr}[\key_j \leftarrow \mathsf{Hyb}_{i-1}(1^\lambda,j) ; c \leftarrow \mathcal{A}(1^\lambda, \key_j) : c = 1]\;-
\]
\[
        \textrm{Pr}[\key_j \leftarrow \mathsf{Hyb}_{i}(1^\lambda,j) ; c \leftarrow \mathcal{A}(1^\lambda, \key_j) : c = 1] \big| < \epsilon_{\textsf{PRG}}  
\]
\end{claim}

\textit{Proof.} Let's fix $i \in \{1, \dots n\}$, $j \in \zerone$. Let $\mathcal{A}$ be a $\textsf{Hyb}$-distinguishing adversary with advantage $\epsilon$ for these values. We use $\mathcal{A}$ to construct a corresponding PRG adversary $\mathcal{B}$. Recall that in the PRG challenge for $G$, the adversary $\mathcal{B}$ is given a value $r$ that is either computed by sampling a seed $s \leftarrow \zerone^\lambda$ and computing $r = G(s)$, or is sampled truly at random $r \leftarrow \zerone^{2(\lambda + n + 2)}$. Algorithm \ref{algo:PRG_challenge} describes the PRG challenge of $\mathcal{B}$ embedded in the $\textsf{Hyb}$-distinguishing challenge of $\mathcal{A}$.

\begin{algorithm*}
    \KwIn{$(1^\lambda, i, j)$}
    \KwInitialisation{Sample random $\alpha \sampled \Z_{2^n}$ \\ 
    Sample random $s^{(1)}_0, s^{(1)}_1 \sampled \zerone^\lambda$ and set $t_{0}^{(1)} = 0,\; t_{1}^{(1)} = 1$} 
    \For{$k = 1..n$}{
        $( (s_j^\mathsf{L} \sep t_j^\mathsf{L}, s_j^\mathsf{R} \sep t_j^\mathsf{R}),
            (\sigma_j^\mathsf{L} \sep \tau_j^\mathsf{L}, \sigma_j^\mathsf{R} \sep \tau_j^\mathsf{R}) ) \leftarrow G(s^{(k)}_j)$ \\
        
        \If{$k < i$}
        {
            $CW^{(k)} \sampled \zerone^{2(\lambda + n + 2)}$
        }
        \Else{
            \lIf{$k = i$}{
                $s^{(i)}_{1-j} \sampled \zerone^\lambda$ and $t^{(i)}_{1-j} = 1 - t^{(i)}_{j}$
            }
            $\big( 
                (s_{1-j}^\mathsf{L} \sep t_{1-j}^\mathsf{L}, s_{1-j}^\mathsf{R} \sep t_{1-j}^\mathsf{R}),
                (\sigma_{1-j}^\mathsf{L} \sep \tau_{1-j}^\mathsf{L}, \sigma_{1-j}^\mathsf{R} \sep \tau_{1-j}^\mathsf{R})
            \big) \leftarrow G(s^{(k)}_{1-j})$ 
            
            \If{$\alpha[k]$}{
                $cw^{(k)} \leftarrow \big( (0^\lambda \sep 0,  s_0^\mathsf{L} \oplus s_1^\mathsf{L} \sep 1),
                (\sigma_0^\mathsf{R} \oplus \sigma_1^\mathsf{R} \sep 1, 0^\lambda \sep 0) \big)$ \\
            }\Else{
                $cw^{(k)} \leftarrow \big( (s_0^\mathsf{R} \oplus s_1^\mathsf{R} \sep 1, 0^\lambda \sep 0), 
                (0^\lambda \sep 0, \sigma_0^\mathsf{L} \oplus \sigma_1^\mathsf{L} \sep 1) \big)$
            }
            \BlankLine
            $CW^{k} \leftarrow cw^{(k)} \oplus G(s^{(k)}_0) \oplus G(s^{(k)}_1)$ \\
            \BlankLine
            $state_{1-j} \leftarrow G(s^{(k)}_{1-j}) \oplus (t_{1-j}^{(k)} \cdot CW^{k}) = ((state_{1-j,0}, state_{1-j,1}),(state'_{1-j,0}, state'_{1-j,1}))$ \\
            Parse $s_{1-j}^{(k+1)} \sep t_{1-j}^{(k+1)} = state_{1-j, \alpha[k]}$
            and $\sigma_{1-j}^{(k+1)} \sep \tau_{1-j}^{(k+1)} = state'_{1-j, 1-\alpha[k]}$
        }
        \BlankLine
        
        $state_{j} \leftarrow G(s^{(k)}_{j}) \oplus (t_{j}^{(k)} \cdot CW^{k}) = ((state_{j,0}, state_{j,1}),(state'_{j,0}, state'_{j,1}))$ \\
        Parse $s_j^{(k+1)} \sep t_j^{(k+1)} = state_{j, \alpha[k]}$
        and $\sigma_j^{(k+1)} \sep \tau_j^{(k+1)} = state'_{j, 1-\alpha[k]}$\\
        
        \If{$k < i$}
        {
            $CW^{k}_{leaf} \sampled \zerone^n$
        }\Else{
            $CW^{k}_{leaf} \leftarrow (-1)^{\tau_1^{(k+1)}} \cdot \left( \sigma_1^{(k+1)} - \sigma_0^{(k+1)} +  \alpha[k] \right) \bmod 2^n$ \textbf{if} $k < i$ \textbf{else} $\zerone^n$
        }
    }
    
    $CW^{(n+1)}_{leaf} \leftarrow (-1)^{t_1^{(n+1)}} \cdot \left( 1 - s_0^{(n+1)} + s_1^{(n+1)} \right) \bmod 2^n$ \textbf{if} $i \le n$ \textbf{else} $\zerone^n$
    
    \KwRet $\kb \leftarrow \shared{\alpha}_j \sep s_j^{(1)} \sep (CW^{(i)})_{i=1..n} \sep (CW_{leaf}^{(i)})_{i=1..n+1}$
\caption{$\mathsf{Hyb}_i$: Hybrid distribution $i$, in which the first $i$ correction words are sampled completely at random, and the remaining correction words are computed honestly.}
\label{algo:hyb}
\end{algorithm*}

\begin{algorithm*}
    \KwIn{$(1^\lambda, (i, j), r)$}
    Sample random $\alpha \sampled \Z_{2^n}$ \\
    Sample  $s^{(1)}_j \sampled \zerone^\lambda$ and set $t_{j}^{(1)} \leftarrow j$ \\
    \For{$k = 1 .. i-1$}{
        $CW^{(k)} \sampled \zerone^{2(\lambda + n + 2)}$\\
        $CW^{(k)}_{leaf} \leftarrow \zerone^n$\\
        $state_j \leftarrow G(s^{(k)}_j) \oplus (t_j^{(k)} \cdot CW^{(k)} = ((state_{j,0}, state_{j,1}),(state'_{j,0}, state'_{j,1}))$ \\
            Parse $s_j^{(k+1)} \sep t_j^{(k+1)} = state_{j, \alpha[k]}$
            and $\sigma_j^{(k+1)} \sep \tau_j^{(k+1)} = state'_{j, 1-\alpha[k]}$\\
        Take $t^{(k+1)}_{1-j} = 1- t^{(k+1)}_{j}$
    }
    
    \BlankLine
    
    $( (s_j^\mathsf{L} \sep t_j^\mathsf{L}, s_j^\mathsf{R} \sep t_j^\mathsf{R}),
            (\sigma_j^\mathsf{L} \sep \tau_j^\mathsf{L}, \sigma_j^\mathsf{R} \sep \tau_j^\mathsf{R}) ) \leftarrow G(s^{(i)}_j)$ \\
    $( (s_{1-j}^\mathsf{L} \sep t_{1-j}^\mathsf{L}, s_{1-j}^\mathsf{R} \sep t_{1-j}^\mathsf{R}), (\sigma_{1-j}^\mathsf{L} \sep \tau_{1-j}^\mathsf{L},\sigma_{1-j}^\mathsf{R} \sep \tau_{1-j}^\mathsf{R}) ) \leftarrow r$ \tcp*{The PRG challenge}
    
    \leIf{$\alpha[i]$}{
        $cw^{(i)} \leftarrow \big( (0^\lambda \sep 0,  s_0^\mathsf{L} \oplus s_1^\mathsf{L} \sep 1),
            (\sigma_0^\mathsf{R} \oplus \sigma_1^\mathsf{R} \sep 1, 0^\lambda \sep 0) \big)$ \\}{
        $cw^{(i)} \leftarrow \big( (s_0^\mathsf{R} \oplus s_1^\mathsf{R} \sep 1, 0^\lambda \sep 0), 
    (0^\lambda \sep 0, \sigma_0^\mathsf{L} \oplus \sigma_1^\mathsf{L} \sep 1) \big)$}
    $\CWi \leftarrow cw^{(i)} \oplus G(s^{(i)}_j) \oplus r$ \\
    
    \For{$x = 0,1$}{
        $state_x \leftarrow G(s^{(i)}_x) \oplus (t_x^{(i)} \cdot \CWi)$ \textbf{if} $x = j$ \textbf{else} $r \oplus (t_x^{(i)} \cdot \CWi)$\label{line:prg_cw} \\
        $state_x = ((state_{x,0}, state_{x,1}),(state'_{x,0}, state'_{x,1}))$ \\
        Parse $s_x^{(i+1)} \sep t_x^{(i+1)} = state_{x, \alpha[i]}$
        and $\sigma_x^{(i+1)} \sep \tau_x^{(i+1)} = state'_{x, 1-\alpha[i]}$
    }
    
    $CW^{(i)}_{leaf} \leftarrow (-1)^{\tau_1^{(i+1)}} \cdot \left( \sigma_1^{(i+1)} - \sigma_0^{(i+1)} +  \alpha[i] \right) \bmod 2^n$
    
    \BlankLine
    \BlankLine
    
    %Compute $(CW^{(k)})_{k>i} \sep (CW_{leaf}^{(k)})_{k>i}$ = $\textsf{RemainingKey}(\alpha, k, s_0^{(i+1)}, s_1^{(i+1)}, t_0^{(i+1)},  t_1^{(i+1)},  \sigma_0^{(i+1)},\sigma_1^{(i+1)}, \tau_0^{(i+1)}, \tau_1^{(i+1)}$) \\
    
    \For{$k = i+1..n$}{
        \For{$x = 0,1$}{
            $( (s_x^\mathsf{L} \sep t_x^\mathsf{L}, s_x^\mathsf{R} \sep t_x^\mathsf{R}),
            (\sigma_x^\mathsf{L} \sep \tau_x^\mathsf{L}, \sigma_x^\mathsf{R} \sep \tau_x^\mathsf{R}) ) \leftarrow G(s^{(k)}_x)$ \\
        }
        \leIf{$\alpha[k]$}{
            $cw^{(k)} \leftarrow \big( (0^\lambda \sep 0,  s_0^\mathsf{L} \oplus s_1^\mathsf{L} \sep 1),
                (\sigma_0^\mathsf{R} \oplus \sigma_1^\mathsf{R} \sep 1, 0^\lambda \sep 0) \big)$ \\}{
            $cw^{(k)} \leftarrow \big( (s_0^\mathsf{R} \oplus s_1^\mathsf{R} \sep 1, 0^\lambda \sep 0), 
        (0^\lambda \sep 0, \sigma_0^\mathsf{L} \oplus \sigma_1^\mathsf{L} \sep 1) \big)$}
        $CW^{(k)} \leftarrow cw^{(k)} \oplus G(s^{(k)}_0) \oplus G(s^{(k)}_1)$ \\
        \For{$x = 0,1$}{
            $state_x \leftarrow G(s^{(k)}_x) \oplus (t_x^{(k)} \cdot CW^{(k)}) = ((state_{x,0}, state_{x,1}),(state'_{x,0}, state'_{x,1}))$ \\
            Parse $s_x^{(k+1)} \sep t_x^{(k+1)} = state_{x, \alpha[k]}$
            and $\sigma_x^{(k+1)} \sep \tau_x^{(k+1)} = state'_{x, 1-\alpha[k]}$
        }
        $CW^{(k)}_{leaf} \leftarrow (-1)^{\tau_1^{(k+1)}} \cdot \left( \sigma_1^{(k+1)} - \sigma_0^{(k+1)} +  \alpha[k] \right) \bmod 2^n$ 
    }
    $CW^{(n+1)}_{leaf} \leftarrow (-1)^{t_1^{(n+1)}} \cdot \left( 1 - s_0^{(n+1)} + s_1^{(n+1)} \right) \bmod 2^n$ \\
        
    \KwRet $\kb = \shared{\alpha}_j \sep s_j^{(1)} \sep (CW^{(i)})_{i=1..n} \sep (CW_{leaf}^{(i)})_{i=1..n+1}$
\caption{PRG Challenge for adversary $\mathcal{B}$}
\label{algo:PRG_challenge}
\end{algorithm*}

Now, consider $\mathcal{B}$’s success in the PRG challenge as a function of $\mathcal{A}$’s success in distinguishing $\mathsf{Hyb}_{i-1}$ from $\mathsf{Hyb}_{i}$. This means that if $\mathcal{A}$ succeeds, then $\mathcal{B}$ will succeeds at its challenge, which implies Claim \ref{claim:hyb_i}.
If, in Algorithm \ref{algo:PRG_challenge}, $r$ is computed \textit{pseudorandomly} using the PRG, then it is clear the generated $\kb$ is distributed as $\mathsf{Hyb}_{i-1}(1^\lambda, j)$.

It remains to show that if $r$ was sampled at random then the generated $\kb$ is distributed as $\mathsf{Hyb}_{i}(1^\lambda, j)$. That is, if $r$ is random, then the corresponding computed values of $s_{1-j}^{(i+1)}$, $CW^{(i)}$ and $CW_{leaf}^{(i)}$ are distributed \textit{randomly} conditioned on the values of $s_j^{(1)} \sep t_j^{(1)} \sep (CW^{(i)})_{i=1..i-1} \sep (CW_{leaf}^{(i)})_{i=1..i-1}$, and the value of $t_{1-j}^{(i)}$ is given by $1-t_{j}^{(i)}$. Note that all remaining values (for $k > i$) are computed as a function of the values computed up to step $i$.

\noindent First, consider $\CWi$, which is computed as such:
\[ \CWi = cw^{(i)} \oplus G(s^{(i)}_j) \oplus r \]

\noindent In particular, when $\alpha[i] = 1$:
$$\begin{array}{l}
    cw^{(i)} \oplus r \\
    = \big( 
        (0^\lambda \sep 0, s_0^\mathsf{L} \oplus s_1^\mathsf{L} \sep 1),
        (\sigma_0^\mathsf{R} \oplus \sigma_1^\mathsf{R} \sep 1, 0^\lambda \sep 0)
    \big) \\
    \oplus \big( 
        (s_{1-j}^\mathsf{L} \sep t_{1-j}^\mathsf{L}, s_{1-j}^\mathsf{R} \sep t_{1-j}^\mathsf{R}), \\
        \qquad (\sigma_{1-j}^\mathsf{L} \sep \tau_{1-j}^\mathsf{L},\sigma_{1-j}^\mathsf{R} \sep \tau_{1-j}^\mathsf{R})
    \big) \\
%\[
%    = \big( 
%        (s_{1-j}^\mathsf{L} \sep  t_{1-j}^\mathsf{L},
%        s_j^\mathsf{L} \oplus s_{1-j}^\mathsf{L} \oplus s_{1-j}^\mathsf{R} %\sep 1 \oplus t_{1-j}^\mathsf{R}),
%\]
%\[
%        (\sigma_j^\mathsf{R} \oplus \sigma_{1-j}^\mathsf{R} \oplus %\sigma_{1-j}^\mathsf{L} \sep 1 \oplus \tau_{1-j}^\mathsf{L},
%        \sigma_{1-j}^\mathsf{R} \sep \tau_{1-j}^\mathsf{R})
%    \big)
%\]
    = \big( 
        (0^\lambda \sep 0, s_j^\mathsf{L} \sep 1),
        (\sigma_j^\mathsf{R} \sep 1, 0^\lambda \sep 0)
    \big) \\
    \oplus \big( 
        (s_{1-j}^\mathsf{L} \sep t_{1-j}^\mathsf{L}, 
        s_{1-j}^\mathsf{L} \oplus s_{1-j}^\mathsf{R} \sep t_{1-j}^\mathsf{R}), \\
        \qquad (\sigma_{1-j}^\mathsf{L} \sep \tau_{1-j}^\mathsf{L},
         \sigma_{1-j}^\mathsf{L} \oplus \sigma_{1-j}^\mathsf{R} \sep \tau_{1-j}^\mathsf{R})
    \big)
    \end{array}$$

\noindent When $\alpha[i] = 0$:
$$\begin{array}{l}
    cw^{(i)} \oplus r \\
    = \big(
        (s_0^\mathsf{R} \oplus s_1^\mathsf{R} \sep 1, 0^\lambda \sep 0), 
        (0^\lambda \sep 0, \sigma_0^\mathsf{L} \oplus \sigma_1^\mathsf{L} \sep 1) 
    \big) \\
    \oplus \big( 
        (s_{1-j}^\mathsf{L} \sep t_{1-j}^\mathsf{L}, s_{1-j}^\mathsf{R} \sep t_{1-j}^\mathsf{R}), \\
        \qquad (\sigma_{1-j}^\mathsf{L} \sep \tau_{1-j}^\mathsf{L},\sigma_{1-j}^\mathsf{R} \sep \tau_{1-j}^\mathsf{R})
    \big) \\
    = \big( 
        (s_j^\mathsf{R} \sep 1, 0^\lambda \sep 0),
        (0^\lambda \sep 0, \sigma_j^\mathsf{L} \sep 1)
    \big) \\
    \oplus \big( 
        (s_{1-j}^\mathsf{L} \oplus s_{1-j}^\mathsf{R} \sep t_{1-j}^\mathsf{L}, 
        s_{1-j}^\mathsf{R} \sep t_{1-j}^\mathsf{R}), \\
        \qquad (\sigma_{1-j}^\mathsf{L}\sep \tau_{1-j}^\mathsf{L},
         \sigma_{1-j}^\mathsf{L} \oplus \sigma_{1-j}^\mathsf{R}   \sep \tau_{1-j}^\mathsf{R})
    \big)
    \end{array}$$

\noindent Independently of the value of $\alpha[i]$, when $r$ is random, the right hand side of the $\oplus$ acts as a perfect one-time pad, and so $\CWi$ is distributed uniformly.

Consider now $CW_{leaf}^{(i)}$, computed as such:

\[ 
 CW^{(i)}_{leaf} \leftarrow (-1)^{\tau_1^{(i+1)}} \cdot \left( \sigma_1^{(i+1)} - \sigma_0^{(i+1)} +  \alpha[i] \right) \bmod 2^n
\]

\noindent Since $\sigma_{1-j}^{(i+1)}$ is distributed randomly conditioned on the previous values computed, it acts as a one-time pad, which ensures that $CW_{leaf}^{(i)}$ is distributed uniformly.

Now, condition on $\CWi$ as well, and consider the value of $s_{1-j}^{(i+1)}$. 
$s_{1-j}^{(i+1)}$ is extracted from $r \oplus t_{1-j}^{(i)} \cdot \CWi$, see Line \ref{line:prg_cw}.
If $t_{1-j}^{(i)} = 0$, $s_{1-j}^{(i+1)}$ is immediately uniformly distributed. If $t_{1-j}^{(i)} = 1$, $r \oplus (t_{1-j}^{(i)} \cdot \CWi) = r \oplus \CWi = r \oplus cw^{(i)} \oplus G(s^{(i)}_j) \oplus r = cw^{(i)} \oplus G(s^{(i)}_j)$. When $\alpha[i] = 1$, the part of $cw^{(i)}$ contributing to $s_{1-j}^{(i+1)}$ is $s_{j}^\mathsf{L} \oplus s_{1-j}^\mathsf{L}$. $s_{1-j}^\mathsf{L}$ is random and hence acts as a perfect one-time pad, so $s_{1-j}^{(i+1)}$ is uniformly distributed. When $\alpha[i] = 0$, the same result is derived using $s_{1-j}^\mathsf{R}$.

\noindent Finally, consider the value of $t_{1-j}^{(i+1)}$. We show that $t_{1-j}^{(i+1)} = 1 - t_{j}^{(i)}$. By construction,
\[
    t_{1-j}^{(i+1)}  =  t_{1-j}^{\textsf{Keep}} \oplus t_{1-j}^{(i)} \cdot t_{CW}^{\textsf{Keep}} 
\]
where $\textsf{Keep} = \textsf{L}$ if  $\alpha[i] = 0$ else $\textsf{R}$, and where $t_{CW}^{\textsf{Keep}} = 1 \oplus t_{0}^{\textsf{Keep}} \oplus t_{1}^{\textsf{Keep}}$.
Further more, by noting that $t_{1-j}^{(i)}$ was set to $1 - t_{j}^{(i)}$,
$$\begin{array}{l}
    t_{j}^{(i+1)} \oplus t_{1-j}^{(i+1)} \\
= (t_{j}^{\textsf{Keep}} \oplus t_{j}^{(i)} \cdot t_{CW}^{\textsf{Keep}} ) \oplus (t_{1-j}^{\textsf{Keep}} \oplus t_{1-j}^{(i)} \cdot t_{CW}^{\textsf{Keep}} ) \\
= t_{j}^{\textsf{Keep}}  \oplus t_{1-j}^{\textsf{Keep}} \oplus (t_{j}^{(i)} \oplus t_{1-j}^{(i)} )\cdot t_{CW}^{\textsf{Keep}} ) \\
= t_{j}^{\textsf{Keep}}  \oplus t_{1-j}^{\textsf{Keep}} \oplus 1 \cdot (1 \oplus t_{0}^{\textsf{Keep}} \oplus t_{1}^{\textsf{Keep}} ) \\
= 1
\end{array}$$
Combining these pieces, we have that in the case of a random PRG challenge $r$, the resulting distribution of $\kb$ as generated by $\mathcal{B}$ is precisely distributed as is $\mathsf{Hyb}_i(1^\lambda, j)$. Thus, the advantage of $\mathcal{B}$ in the PRG challenge experiment is equivalent to the advantage $\epsilon$ of $\mathcal{A}$ in distinguishing $\mathsf{Hyb}_{i-1}(1^\lambda, j)$ from $\mathsf{Hyb}_i(1^\lambda, j)$. The runtime of $\mathcal{B}$ is equal to the runtime of $\mathcal{A}$ plus a fixed polynomial $p'(\lambda)$. Thus for any $T'  \le T - p'(\lambda)$, it must be that the distinguishing advantage $\epsilon$ of $\mathcal{A}$ is bounded by $\epsilon_{\mathsf{PRG}}$, which concludes the proof of Claim \ref{claim:hyb_i}.

Combining all the steps, for $i \in \{1, \dots n\}$, we have thus proven that $\mathsf{Hyb}_0(1^\lambda, j)$ and $\mathsf{Hyb}_n(1^\lambda, j)$ are computationally indistinguishable for any adversary, for $j\in\zerone$.
On the other hand, we can also prove:
\begin{claim}\label{claim:hyb_n}
\[ \mathsf{Hyb}_{n}(1^\lambda,j) = \mathsf{Hyb}_{n+1}(1^\lambda,j) \]
\end{claim}

\textit{Proof.} In $\mathsf{Hyb}_{n+1}$, $CW_{leaf}^{(n+1)} \sampled \zerone^n$. In $\mathsf{Hyb}_{n}$, $CW_{leaf}^{(n+1)} = (-1)^{t_1^{(n+1)}} \cdot ( 1 - s_0^{(n+1)} + s_1^{(n+1)} ) \bmod 2^n$, with $s_{1-j}^{(n+1)}$ distributed randomly conditioned on the previous values computed. $ s_{1-j}^{(n+1)}$ acts as a one-time pad which perfectly hides other values. Hence, $CW_{leaf}^{(n+1)}$ is also uniformly distributed in this case.

This concludes the proof of security of our FSS comparison protocol.

\section{Encoding Precision}\label{appendix:precision}
We have studied the impact of lowering the encoding space of the input to our function secret sharing protocol from $\Z_{2^{32}}$ to $\Z_{2^{k}}$ with $k$ < 32. Finding the lowest $k$ guaranteeing good performance is an interesting challenge as the function keys size is directly proportional to it. This has to be done together with reducing fixed precision from 3 decimals down to 1 decimal to ensure private values aren't too big, which would result in higher failure rate in our private comparison protocol.
We have reported in Table \ref{table:precision} our findings on Network-1, which is pre-trained and then evaluated in a private fashion. 

\begin{table}[h!]
\begin{center}
\begin{tabular}{c c  c c c c c}
 Decimals & $\Z_{2^{12}}$ & $\Z_{2^{16}}$ & $\Z_{2^{20}}$ & $\Z_{2^{24}}$ & $\Z_{2^{28}}$ & $\Z_{2^{32}}$ \\ 
\hline
 1 & - & - & - & - & - & 9.5 \\ 
 2 & 69.4 & 96.0 & 97.9 & 98.1 & 98.0 & 98.1 \\
 3 & 10.4 & 76.2 & 96.9 & 98.1 & 98.2 & 98.1 \\
 4 & 9.7 & 14.3  & 83.5 & 97.4 & 98.1 & 98.2 \\
\end{tabular}
\vspace{1mm}
\caption{Accuracy (in \%) of Network-1 given different precision and encoding spaces}
\label{table:precision}
\end{center}
\end{table}

What we observe is that 3 decimals of precision is the most appropriate setting to have an optimal precision while allowing to slightly reduce the encoding space down to $\Z_{2^{24}}$ or $\Z_{2^{28}}$. Because this is not a massive gain and in order to keep the failure rate in comparison very low, we have kept $\Z_{2^{32}}$ for all our experiments.

\section{Implementation Details}\label{appendix:implem}
\subsection{Pseudo-Random Generator}\label{app:prg}

The PRG is implemented using a Matyas-Meyer-Oseas one-way compression function as in \cite{wang2017splinter}, with an AES block cipher. We concatenate several fixed key block ciphers to achieve the desired output length: $G(x) = E_{k_1}(x) \oplus x \sep E_{k_2}(x) \oplus x \sep \dots$. Those keys are fixed and hard-coded. We set $\lambda = 127$ to be able to use only 2 blocks for equality and 4 blocks for comparison. Note that for comparison we would theoretically only need 3 blocks, although our current implementation uses 4.

\subsection{Unrolling Convolutions}\label{app:unrolling}

\begin{figure}[htb!]
  \centering
  \includegraphics[width=.6\linewidth]{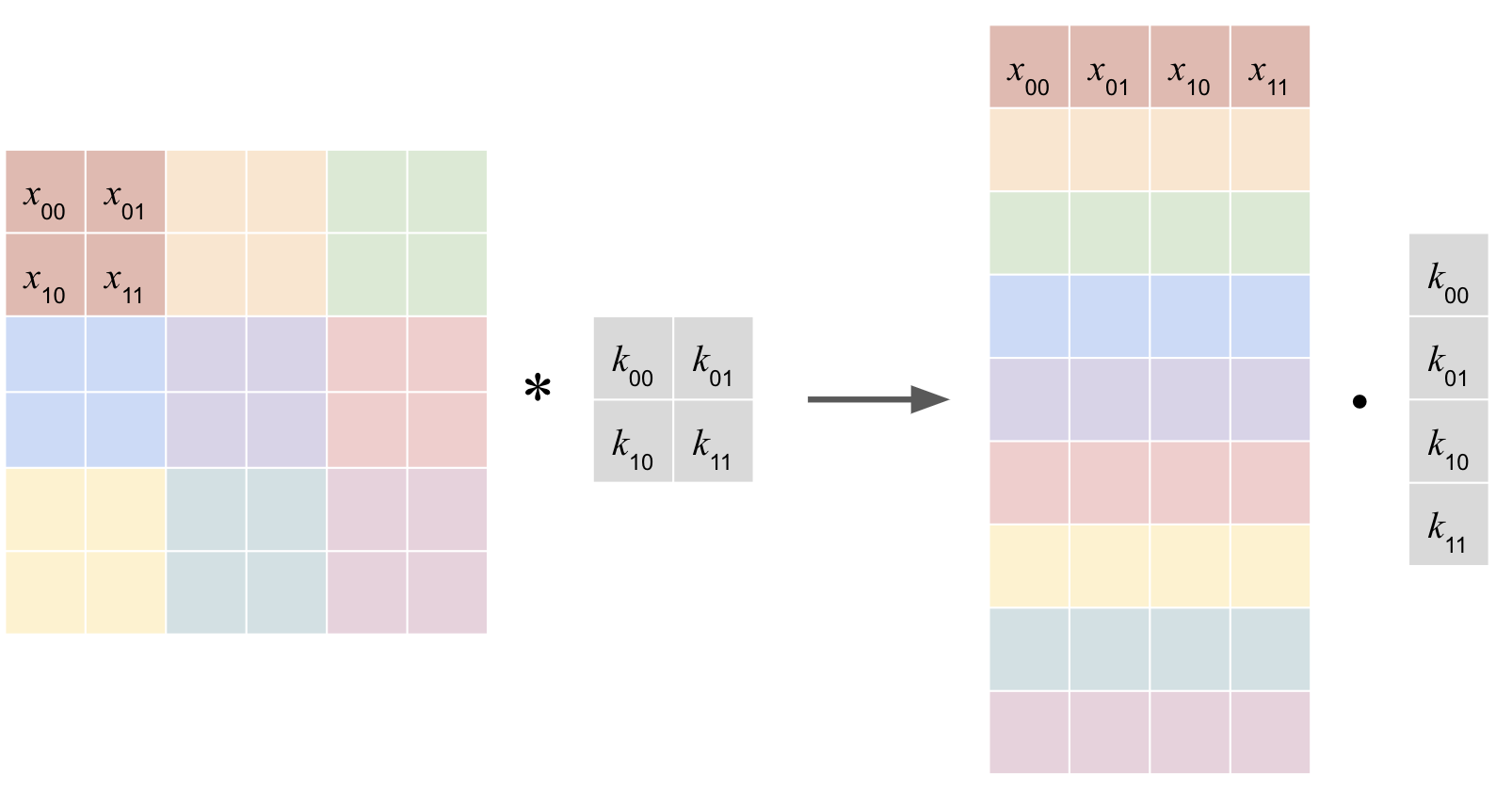}
  \caption{Illustration of unrolling a convolution with kernel size $k = 2$ and stride $s = 2$. }\label{fig:unrolling}
\end{figure}

Figure \ref{fig:unrolling} illustrates how to transform a convolution operation into a single matrix multiplication.

\subsection{MaxPool and Optimisation}\label{app:maxpool}

\begin{figure}[htb!]
  \centering
  \includegraphics[width=0.9\linewidth]{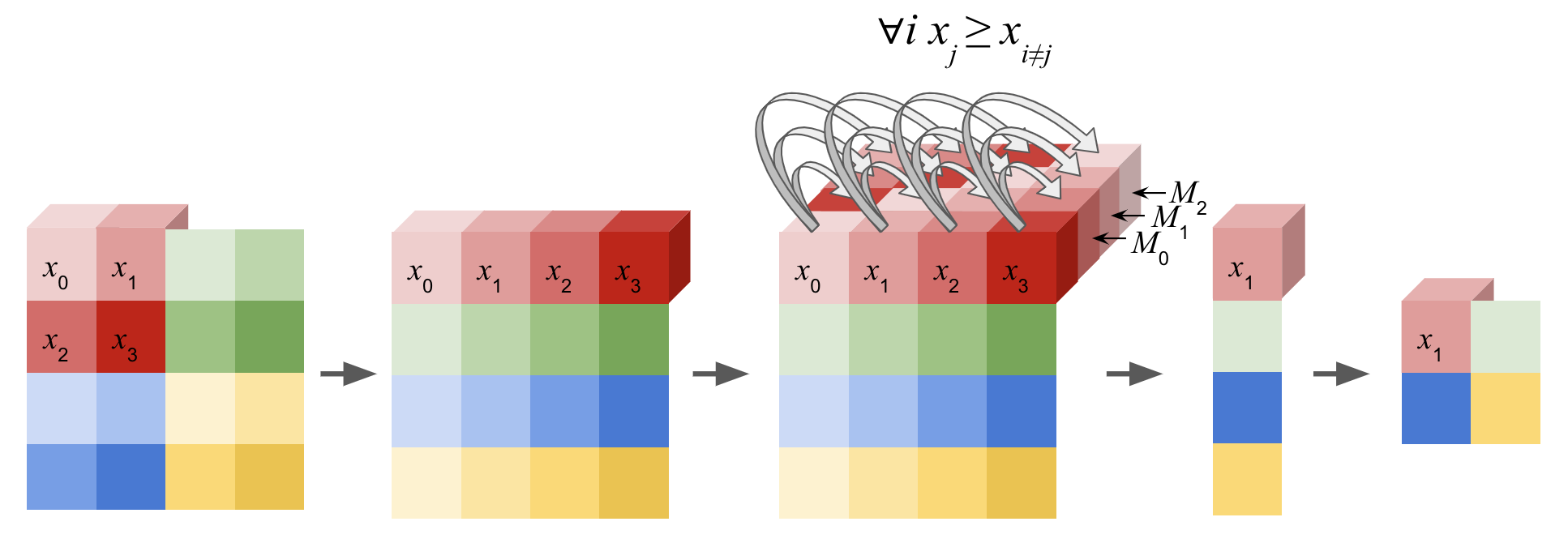}
  \caption{Illustration of MaxPool with kernel size $k = 2$ and stride $s = 2$. }\label{fig:maxpool}
\end{figure}

Figure \ref{fig:maxpool} illustrates how MaxPool uses ideas from matrix unrolling and argmax computation. Notations present in the figure are consistent with the explanation of argmax using pairwise comparison in Section \ref{sec:common_ml_op}. The $m \times m$ matrix is first unrolled to a roughly $(m/s)^2 \times k^2$ matrix. It is then expanded on $k^2$ layers, each of which is shifted by a step of 1. Next, $(m/s)^2k^2(k^2-1)$ pairwise comparisons are then applied simultaneously between the first layer and the other ones, and for each $x_i$ we sum the result of its $k-1$ comparison and check if it equals $k-1$. We multiply this boolean by $x_i$ and sum up along a line (like $x_0$ to $x_3$ in the figure). Last, we restructure the matrix back to its initial structure.

In addition, when the kernel size $k$ is 2, rows are only of length 4 and it can be more efficient to use a binary tree approach instead, i.e. compute the maximum of columns 0 and 1, 2 and 3 and the max of the result: it requires $2\log_2(k^2) = 4$ rounds of communication but only approximately $\ (k^2-1)(m/s)^2$ private comparisons, compared to a fixed 3 rounds and approximately $k^4 (m/s)^2$. We found in practice that this $4\times$ speed-up factor in computation is worth an additional communication round.

Interestingly, average pooling can be computed locally on the shares without interaction because it only includes mean operations, but we didn't replace MaxPool operations with average pooling to avoid distorting existing neural networks architecture.

\subsection{Breaking Ties in Argmax}\label{app:argmax}

Algorithm \ref{algo:breakties} provides a probabilistic method to break ties in the argmax output, which can be used on secret shared input as well.

\begin{algorithm}[t]
    \KwIn{$\vec \delta = (\delta_1, \dots, \delta_m$), with $\forall i, \delta_i \in \zerone$}
    \KwOut{$\delta_j$ with $j \sampled \{i, \delta_i = 1\}$}
    $\vec x = \mathsf{CumSum}(\vec \delta)$ \\ 
    $r \sampled \left [0, \vec x[-1] \right [$ \\
    $\vec c = \vec x > r$ \\
    Compute $\vec c_{\gg}$ by shifting $\vec c$  by one number on the right and padding with 0 \\
    \KwRet $\vec c - \vec c_{\gg}$
\caption{$\mathsf{BreakTies}$ algorithm to guarantee one-hot output}
\label{algo:breakties}
\end{algorithm}

\subsection{BatchNorm Approximation}\label{app:batchnorm}

\begin{figure}[htb!]
  \centering
  \includegraphics[width=\linewidth]{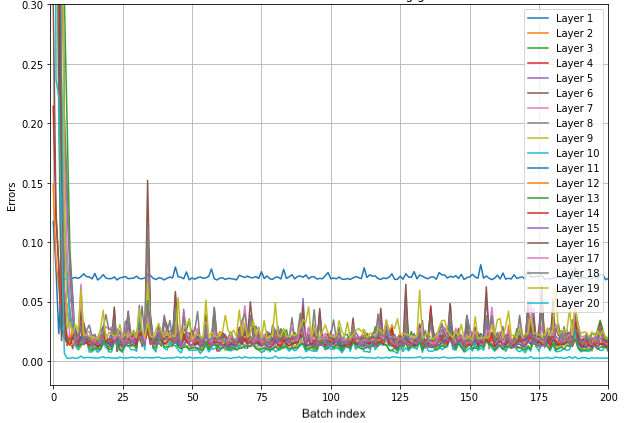}
  \caption{BatchNorm relative error as training goes on}\label{fig:bn_time}
\end{figure}

\begin{figure}[htb!]
  \centering
  \includegraphics[width=0.7\linewidth]{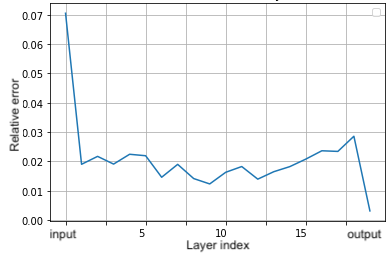}
  \caption{Relative error of the BatchNorm layers across the model}\label{fig:bn_layer}
\end{figure}

The BatchNorm layer is the only one in our implementation which requires a polynomial approximation during training. We have therefore experimented how this approximation can alter the behaviour of a deep network such as ResNet18 on a simple dataset like the hymenoptera dataset\footnote{\url{https://download.pytorch.org/tutorial/hymenoptera_data.zip}}. We follow the PyTorch transfer learning tutorial\footnote{\url{https://pytorch.org/tutorials/beginner/transfer_learning_tutorial.html}}
and use a pretrained version of ResNet18. We retrain all the layers for 25 epochs, but we replace the BatchNorm layers with our approximated version which behaves as such: as a general rule the inverse for the variance is computed using 3 iterations of the Newton methods and we use the result of the previous batch as an initial approximation. For the first batch of the epoch where no approximation is available, we use instead 50 iterations. For the first BatchNorm layer, since the variance can change significantly because of the input diversity, we systematically use more iterations, up to 60. For the last layer, as the variance is very small, its inverse can have a large amplitude. Therefore we don't use the result of the previous batch and perform 10 iterations instead. We report in Figure \ref{fig:bn_time} the evolution of the error as the model train on more batches. As one can see, the error dramatically shrinks after 10 batches, which shows how beneficial it is to use previous batch computations. It almost stays below $5\%$, except for the error of the first layer, for which we would need even more iterations to have significant improvements. The average error per layer is reported in Figure \ref{fig:bn_layer}, and is around $2\%$.

This is of course an experimental result, but it shows that the BatchNorm approximation can be tailored for specific models to allow for efficient training with little round overhead. Table \ref{table:batchnorm} shows that we can indeed achieve a high accuracy, especially using the initial guess of the previous batch, although it remains below the accuracy of a model trained with an exact BatchNorm.

\section{Extended Results about Private Inference}\label{appendix:extended_inference}
We provide additional results about our inference experiment in Table \ref{table:extended_inference}.

\begin{table*}[htb]
\caption{Comparison of the inference time between secure frameworks on several popular neural network architectures. For each network we report in order the batch size used and the size of the preprocessing material in MB for the CPU setting.\label{table:extended_inference}}
\vspace*{-.111cm}
\begin{center}
\begin{tabular}{ l r | c c c | c c c | c c c }
\hline
  & & \multicolumn{3}{c|}{Network-1} & \multicolumn{3}{c|}{Network-2} & \multicolumn{3}{c}{LeNet}\\ 
  & & \multicolumn{2}{c}{Batch Size} & Preprocessing & \multicolumn{2}{c}{Batch Size} & Preprocessing & \multicolumn{2}{c}{Batch Size} & Preprocessing \\
Framework & \!\!\!\!\!\!Dataset\!\!
    & CPU & GPU & Comm. (MB)
    & CPU & GPU & Comm. (MB) 
    & CPU & GPU & Comm. (MB) \\
\hline
AriaNN    & \!\!\!\!\!\!\!\!MNIST\!\! 
    & 128 & 128 & 0.36
    & 128 & 128 & 9.98
    & 128 & 128 & 14.7 \\

\hline
\end{tabular}

\vspace*{0.1cm}

\begin{tabular}{ l r | c c c | c c c | c c c }
\hline
  & & \multicolumn{3}{c|}{AlexNet} & \multicolumn{3}{c|}{VGG16} & \multicolumn{3}{c}{ResNet18}\\ 
  & & \multicolumn{2}{c}{Batch Size} & Preprocessing & \multicolumn{2}{c}{Batch Size} & Preprocessing & \multicolumn{2}{c}{Batch Size} & Preprocessing \\
Framework & Dataset  
    & CPU & GPU & Comm. (MB)
    & CPU & GPU & Comm. (MB) 
    & CPU & GPU & Comm. (MB) \\
\hline
AriaNN & CIFAR-10\!\!
    & 128 & 128 & 24.6
    & 64 & 14 & 277
    & -   & -   & - \\
AriaNN & $\!\!\!\!\!\!\!\!\small{64\!\times\!64}$ ImageNet\!\!
    & 128 & 64 & 88.4 
    & 16  & 3 & 1124
    & -   & -   & - \\
AriaNN & $\!\!\!\!\!\!\!\!\!\!\!\!\!\!\!\small{224\!\times\!224}$ Hymenoptera\!\!
    & -   & -   & - 
    & -   & -   & - 
    & 8   & 1   & 3254 \\
                 
\hline
\end{tabular}

%% CPU
% Resnet18: BS = 8
% VGG16 CIFAR10: 64
% VGG16 Tiny Imagenet: 16 (all 32 -> 7.1s ; all 64 -> 6.48)

%% GPU
% Alexnet Tiny Imagenet BS = 64
% VGG16 CIFAR10: batch_size = 14
% VGG16 Tiny Imagenet: BS = 3 
% ResNet18 GPU BS = 1

\vspace*{-.322cm}
\end{center}
\end{table*}

\section{Datasets and Networks Architecture}
\subsection{Datasets}\label{app:datasets}
% \!\\
\textit{This section is taken almost verbatim from \cite{wagh2020falcon}.}

We select 4 datasets popularly used for training image
classification models: MNIST \cite{lecun2010mnist}, CIFAR-10 \cite{krizhevsky2014cifar},
64$\times$64 Tiny Imagenet \cite{wu2017tiny} and Hymenoptera, a subset of the Imagenet dataset \cite{imagenet_cvpr09} composed 224$\times$224 pixel images.

\textbf{MNIST} 
MNIST \cite{lecun2010mnist} is a collection of handwritten
digits dataset. It consists of 60,000 images in the training set and 10,000 in the test set. Each image is a
28$\times$28 pixel image of a handwritten digit along wit
a label between 0 and 9. We evaluate Network-1, Network-2, and the LeNet network on this dataset.

\textbf{CIFAR-10} 
CIFAR-10 \cite{krizhevsky2014cifar} consists of 50,000 images in the training set and 10,000 in the test set. It is composed of 10
different classes (such as airplanes, dogs, horses, etc.) and there are 6,000 images of each class with each image
consisting of a colored 32$\times$32 image. We perform
private training of AlexNet and inference of VGG16
on this dataset.

\textbf{Tiny ImageNet} 
Tiny ImageNet \cite{wu2017tiny} consists of two datasets
of 100,000 training samples and 10,000 test samples
with 200 different classes. The first dataset is composed of colored 64$\times$64 images and we use it with AlexNet and VGG16. The second is composed of colored 224$\times$224 images and is used with ResNet18.

\textbf{Hymenoptera}
Hymenoptera is a dataset extracted from the ImageNet database. It is composed of 245 training and 153 test colored 224$\times$224 images, and was first proposed as a transfer learning task.

\subsection{Model Description}\label{appendix:NNarchi}
\!

We have selected 6 models for our experimentations. Description on the first 5 models is taken verbatim from \cite{wagh2020falcon}.

\textbf{Network-1} 
A 3-layered fully-connected network with ReLU used in SecureML \cite{mohassel2017secureml}.
    
\textbf{Network-2} 
A 4-layered network selected in MiniONN \cite{liu2017oblivious} with 2 convolutional and 2 fully-connected layers, which uses MaxPool in addition to ReLU activation.
    
\textbf{LeNet}
This network, first proposed by LeCun et al. \cite{lecun1998gradient}, was used in automated detection of zip codes and digit recognition. The network contains 2 convolutional layers and 2 fully connected layers.
    
\textbf{AlexNet}
AlexNet is the famous winner of the 2012 ImageNet ILSVRC-2012 competition \cite{krizhevsky2012alexnet}. It has 5 convolutional layers and 3 fully connected layers and it can use batch normalization layers for stability and efficient training.
    
\textbf{VGG16}
VGG16 is the runner-up of the ILSVRC-2014 competition \cite{simonyan2014vgg}. VGG16 has 16 layers and has about 138M parameters.
    
\textbf{ResNet18}
ResNet18 \cite{he2016resnet} is the runner-up of the ILSVRC-2015 competition. It is a convolutional neural network that is 18 layers deep, and has 11.7M parameters. It uses batch normalisation and we are the first private deep learning framework to evaluate this network.

\subsection{Models Architecture}\label{appendix:NNarchi_detailed}

Unless otherwise specified, our models follow their standard architecture as provided by the \verb|torchvision| library (version 0.5), except for smaller models such as Network-1, Network-2 and LeNet which are respectively detailed in \cite{mohassel2017secureml}, \cite{liu2017oblivious} and \cite{lecun1998gradient}.

For the CIFAR-10 version of AlexNet, we follow the architecture of \cite{wagh2020falcon} which includes BatchNorm layers and is available on the \href{https://github.com/snwagh/falcon-public/blob/b0d209f64b68f28374b40ab5c6239db185b10d14/src/secondary.cpp#L1041}{FALCON GitHub}. For the $64\!\times\!64$ Tiny Imagenet version of AlexNet however, we used the standard architecture from PyTorch since it allows us to have a pretrained network and the version of FALCON seemed non-standard. 

We have adapted the classifier parts of AlexNet, VGG16 and ResNet18 to the different datasets we use.
\begin{itemize}

    \item For AlexNet on Tiny Imagenet, we use 3 fully connected layers with respectively 1024, 1024 and 200 neurons, and ReLU activations between them.
    \item For VGG16  we use 3 fully connected layers with respectively 4096, 4096 and 10 or  200 neurons for the last layer depending on the dataset used. We also use ReLU activations between them.
    \item For ResNet18, we use a single fully connected layer to map the 512 output logits to the appropriate number of classes.
\end{itemize}

Note also that we permute ReLU and Maxpool where applicable like in \cite{wagh2020falcon}, as this is strictly equivalent in terms of output for the network and reduces the number of comparisons. More generally, we don't proceed to any alteration of the network behaviour except with the approximation on BatchNorm. This improves the usability of our framework as it allows us to use pre-trained neural networks from standard deep learning libraries like PyTorch and to encrypt them with a single line of code.

\end{document}